%% file: recent_advances_in_3d_object_.tex
\pdfoutput=1

\documentclass[a4paper,11pt, notitlepage]{article}

\usepackage{pslatex}	    %
\usepackage[usenames,dvipsnames]{color}
\usepackage[latin1]{inputenc}
\usepackage[T1]{fontenc}
\usepackage{hyperref}
\usepackage{amsmath}
\usepackage{gensymb}
\usepackage{multirow}
\usepackage{colortbl}
\usepackage{xspace}
\usepackage{booktabs}
\usepackage{array}
\usepackage{graphicx}
\usepackage{named}
\usepackage{xspace}
\usepackage{pdfpages}
\usepackage[table,xcdraw]{xcolor}
\usepackage{siunitx}
\usepackage{eurosym}
\usepackage{float}
\usepackage{lipsum}

\newcommand{\ie}{\textit{i.e.}}

\input{defs}

\begin{document}

\title{Recent Advances in 3D Object and Hand Pose Estimation}
\author{Vincent Lepetit\\
  \href{mailto:vincent.lepetit@enpc.fr}{vincent.lepetit@enpc.fr}\\
LIGM, Ecole des Ponts, Univ Gustave Eiffel, CNRS, Marne-la-Vall\'ee, France}
\date{}
\maketitle

\begin{abstract}
3D object and hand pose estimation have huge potentials for Augmented Reality, to enable tangible interfaces, natural interfaces, and blurring the boundaries between the real and virtual worlds.  In this chapter, we present the recent developments for 3D object and hand pose estimation using cameras, and discuss their abilities and limitations and the possible future development of the field.
\end{abstract}

\section{3D Object and Hand Pose Estimation for Augmented Reality}

\input{intro}

\section{Formalization}
\input{formalization}

\section{Challenges of 3D Pose Estimation using Computer Vision}
\input{challenges}

\section{Early Approaches to 3D Pose Estimation and Their Limits}
\input{early_approaches}

\section{Machine Learning and Deep Learning}
\label{sec:ml_dl}
\input{ml_dl}

\section{Datasets}
\label{sec:datasets}

Datasets have  become an important aspect  of 3D pose estimation,  for training,
evaluating,  and comparing  methods.  We  describe  some for  object, hand,  and
hand+object pose estimation below.

\subsection{Datasets for Object Pose Estimation}
\label{sub:object_datasets}
\label{sec:object_datasets}
\input{object_datasets}

\subsection{Datasets for Hand Pose Estimation}

\input{hand_datasets}

\subsection{Datasets for Object and Hand Pose Estimation}
\label{sec:hand_object_datasets}
\input{hand_object_datasets}

\subsection{Metrics}
\label{sec:metrics}
\input{metrics}

\section{Modern Approaches to 3D Object Pose Estimation}
\label{sec:object_pose}

\input{object_pose}

\section{3D Pose Estimation for Object Category}
\label{sec:category}
\input{category_pose}

\input{hand_pose}

\section{3D Object+Hand Pose Estimation}
\input{object_hand_pose}

\section{The Future of 3D Object and Hand Pose Estimation}
\input{future}

\bibliographystyle{named}
\bibliography{string,recent_advances_in_3d_object_}

\end{document}

%% file: defs.tex
\newcommand{\bM}{{\bf M}}

\newcommand{\bR}{{\bf R}}

\newcommand{\be}{{\bf e}}

\newcommand{\bm}{{\bf m}}
\newcommand{\bt}{{\bf t}}

\newcommand{\calL}{\mathcal{L}}

\newcommand{\calT}{\mathcal{T}}
\newcommand{\calV}{\mathcal{V}}

\newcommand{\ADD}{\text{ADD}}

\newcommand{\RigidMotion}{\text{Tr}}
\newcommand{\pose}{{\bf p}}
\newcommand{\Proj}{\text{Proj}}

%% file: intro.tex
\begin{figure}
  \begin{center}
    \begin{tabular}{cc}
      \includegraphics[width=0.4\linewidth]{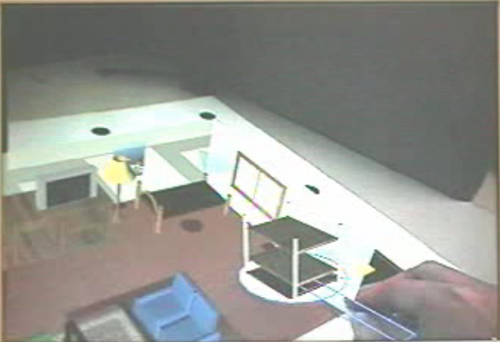} &
      \includegraphics[width=0.4\linewidth]{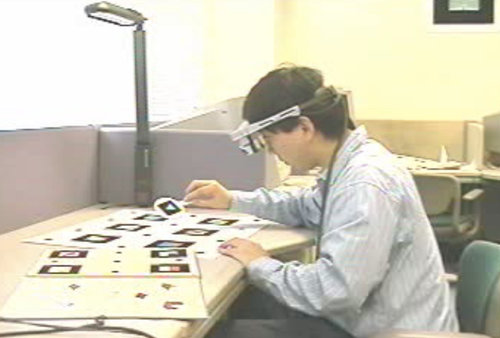} \\
    \end{tabular}
  \end{center}
  \caption{\label{fig:magic_book} Augmented  Reality~(left) and external~(right)
    views of  the early  Magic book  developed by  Kazo and  Billinghurst.  This
    visionary experiment shows the importance  of real object manipulation in AR
    applications.}
\end{figure}

An  Augmented Reality  system  should not  be limited  to  the visualization  of
virtual elements integrated to the real world, it should also \textit{perceive}
the user and the  real  world  surrounding  the   user.   As  even  early  Augmented  Reality
applications demonstrated, this is required to provide rich user experience, and
unlock new possibilities.  For example, the magic book application developed by
Billinghurst     and    Kazo~\cite{Billinghurst01}     and    illustrated     by
Figure~\ref{fig:magic_book}, shows  the importance  of being able  to manipulate
real objects in  Augmented Reality. It featured a real  book, from which virtual
objects would pop up, and these virtual objects could be manipulated by the user
using a small real ``shovel''.

This early  system relied  on visual markers  on the real  book and  the shovel.
These markers would become the core of the popular ARToolkit, and were essential
to robustly estimate the locations and  orientations of objects in the 3D space,
a  geometric information  required for  the  proper integration  of the  virtual
objects with the real book, and their manipulation by the shovel.

Visual markers, however, require a modification  of the real world, which is not
always possible nor desirable in practice. Similarly, magnetic sensors have also
been  used to  perceive  the spatial  positions  of real  objects,  in order  to
integrate them to  the augmented world.  However, they share  the same drawbacks
as  visual  markers.

Being able to perceive the user's hands is also very important, as the hands can
act as an interface between the  user's intention and the virtual world.  Haptic
gloves  or various  types of  joysticks  have been  used to  capture the  hands'
locations and motions.  In particular, input pads are still popular with current
augmented and virtual reality platforms,  as they ensure robustness and accuracy
for the perception of the user's actions.  Entirely getting rid of such hardware
is extremely desirable, as it  makes Augmented Reality user interfaces intuitive
and natural.

Researchers have thus  been developing Computer Vision  approaches to perceiving
the real  world using  simple cameras  and without having  to engineer  the real
objects with  markers or sensors,  or the user's  hands with gloves  or handles.
Such  perception problem  also  appears  in fields  such  as  robotics, and  the
scientific literature on this topic is extremely rich.  The goal of this chapter
is to introduce the reader to 3D object and hand perception based on cameras for
Augmented Reality applications,  and to the scientific literature  on the topic,
with a focus on the most recent techniques.

In  the following,  we will  first  detail the  problem of  3D pose  estimation,
explain why it is difficult, and review  early approaches to motivate the use of
Machine Learning  techniques. After  a brief introduction  to Deep  Learning, we
will discuss the literature on 3D  pose estimation for objects, hands, and hands
manipulating objects through representative works.

%% file: formalization.tex
\begin{figure}
  \begin{center}
    \begin{tabular}{c|c}
      \includegraphics[width=0.48\linewidth]{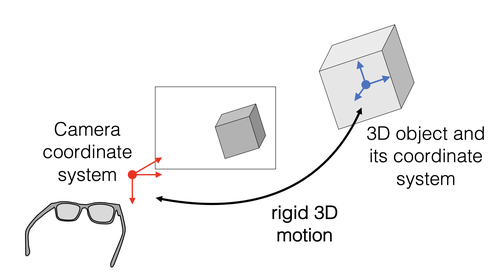} &
      \includegraphics[width=0.48\linewidth]{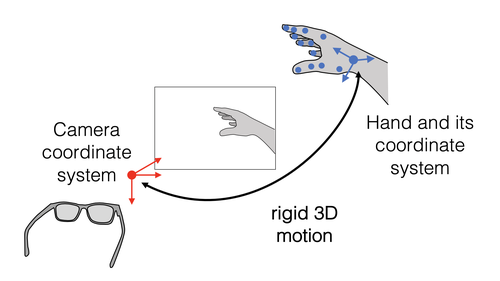} \\
      (a) & (b) \\
    \end{tabular}
  \end{center}
  \caption{\label{fig:poses} (a) The 3D pose of a rigid object can be defined as
    a 3D  rigid motion between  a coordinate system  attached to the  object and
    another coordinate system, for example  one attached to an Augmented Reality
    headset. (b)  One way to define  the 3D pose of  a hand can be  defined as a
    rigid motion between a coordinate system  attached to one of the joints (for
    example the wrist)  and another coordinate system, plus the  3D locations of
    the joints in  the first coordinate system. 3D rigid  motions can be defined
    as the composition of a 3D rotation and a 3D translation. }
\end{figure}

In the rest of this chapter, we will define the 3D pose of a rigid object as its
3D  location   and  orientation   in  some  coordinate   system,  as   shown  in
Figure~\ref{fig:poses}. In the case of Augmented Reality, this coordinate system
should  be directly  related to  the headset,  or to  the tablet  or smartphone,
depending on the visualization system, \textit{and}  to the object. This is thus
different from Simultaneous  Localization and Mapping~(SLAM), where  the 3D pose
of the camera can  be estimated in an arbitrary coordinate system  as long as it
is consistent over time.

In  general,  this  3D pose  has  thus  6  degrees  of  freedom: 3  for  the  3D
translation, and 3 for the 3D  orientation, which can be represented for example
with  a 3D  rotation matrix  or a  quaternion.  Because  of these  6 degrees  of
freedom, the 3D pose  is also called '6D pose' in the  literature, the terms '3D
pose' and '6D pose' thus refer to the same notion.

Articulated objects, such  as scissors, or even more deformable  objects such as
clothes   or   sheets   of   papers   have   also   been   considered   in   the
literature~\cite{Salzmann10b,Pumarola18}. Their positions and shapes in space have
many more degrees  of freedom, and specific representations of  these poses have
been  developed.  Estimating  these  values from  images  robustly remains  very
challenging.

The 3D  pose of  the hand  is also  very complex,  since we  would like  to also
consider the  positions of  the individual  fingers.  This  is by  contrast with
gesture recognition  for example,  which aims at  assigning a  distinctive label
such as  'open' or 'close' to  a certain hand  pose.  Instead, we would  like to
estimate  continuous  values in  the  3D  space.   One  option among  others  to
represent the 3D pose of a hand is to consider the 3D locations of each joint in
some coordinate system.  Given a hand model with bone length and rotation angles
for all  DoF of the  joints, forward  kinematics~\cite{Gustus12} can be  used to
calculate     the     3D     joint     locations.      Reciprocally,     inverse
kinematics~\cite{Tolani00} can be applied to obtain joint angles and bone length
from the  3D joint locations.   In addition, one may  also want to  estimate the
shapes of the user's hands, such as their sizes or the thickness of the fingers.

To be useful for Augmented Reality,  3D pose estimation has to run continuously,
in real-time.  When using computer vision, it  means that it should be done from
a single image, or stereo images captured at the same time, or RGB-D images that
provide both color and depth data, maybe  also relying on the previous images to
guide or stabilize the 3D poses  estimations.  Cameras with large fields of view
help, as they  limit the risk of the  target objects to leave the  field of view
compared  to more  standard cameras.

Stereo camera  rigs, made  of two  or more  cameras, provide  additional spatial
information  that can  be exploited  for estimating  the 3D  pose, but  make the
device more  complex and  more expensive.   RGB-D cameras  are currently  a good
trade-off, as they also provide, in addition to a color image, 3D information in
the form of a depth  map, \ie a depth value for each pixel,  or at least most of
the pixels of the color image.  ``Structured light'' RGB-D cameras measure depth
by projecting  a known pattern  in infrared  light and capturing  the projection
with an  infrared camera.  Depth can  then be estimated from  the deformation of
the pattern.  ``Time-of-flight''  cameras are based on pulsed  light sources and
measure the time a light pulse takes to travel from the emitter to the scene and
come back after reflection.

With  structured-light technology,  depth  data  can be  missing  at some  image
locations, especially along  the silhouettes of objects, or on  dark or specular
regions.  This technology also struggles to work outdoor, because of the ambient
infrared sunlight.   Time-of-flight cameras  is also  disturbed outdoor,  as the
high  intensity of  sunlight causes  a quick  saturation of  the sensor  pixels.
Multiple  reflections   produced  by   concave  shapes   can  also   affect  the
time-of-flight.

But  despite  these drawbacks,  RGB-D  cameras  remain  however very  useful  in
practice when  they can be  used, and algorithms  using depth maps  perform much
better than algorithms  relying only on color information---even  though the gap
is decreasing in recent research.

%% file: challenges.tex
\begin{figure}
  \begin{center}
    \begin{tabular}{ccc}
      \fbox{\includegraphics[height=0.2\linewidth]{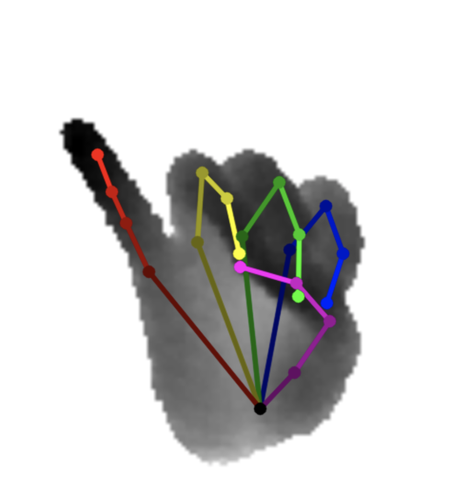}} &
      \includegraphics[height=0.2\linewidth]{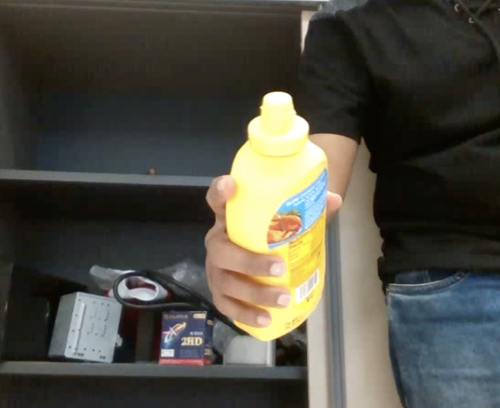} &
      \fbox{\includegraphics[height=0.2\linewidth]{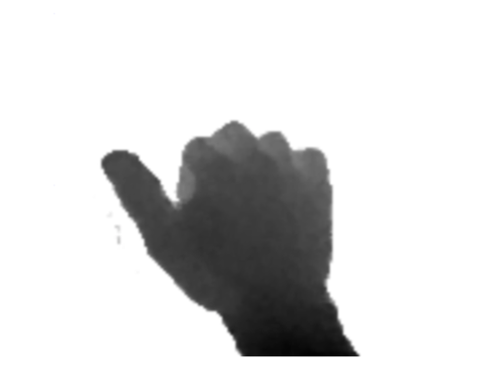}}\\
      (a) & (b) & (c)\\
    \end{tabular}
    \begin{tabular}{cc}
      \includegraphics[height=0.18\linewidth]{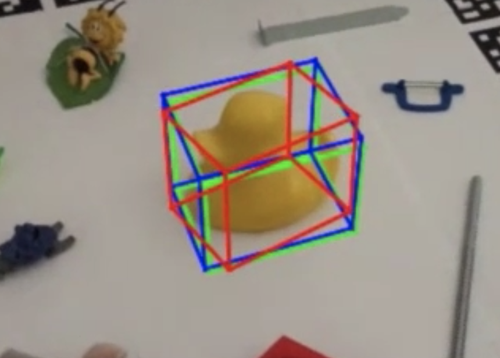}
      \includegraphics[height=0.18\linewidth]{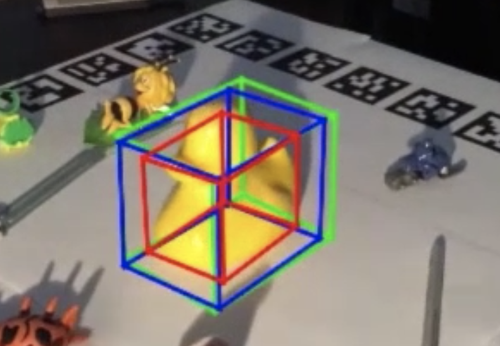} &
      \includegraphics[height=0.18\linewidth]{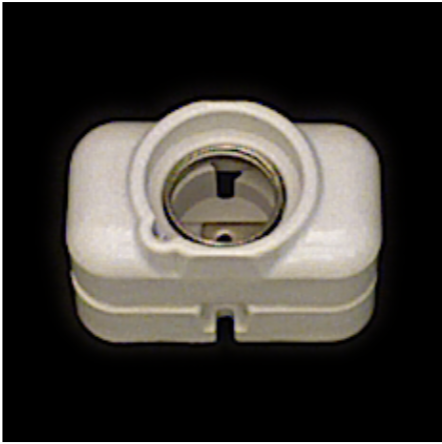}
      \includegraphics[height=0.18\linewidth]{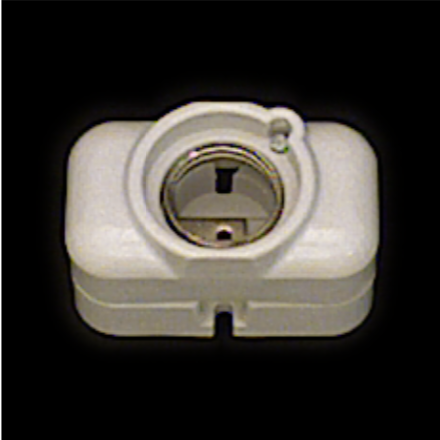} \\
      (d) & (e)
    \end{tabular}
  \end{center}
  \caption{\label{fig:challenges} Some  challenges when estimating the  3D poses
    of  objects and  hands. (a)  Degrees of  freedom. The  human hand  is highly
    articulated, and many parameters have to be estimated to correctly represent
    its pose.   (b) Occlusions.  In this  case, the hand partially  occludes the
    object, and the object partially  occludes the hand. (c) Self-occlusions can
    also  occur in  the  case of  hands, especially  in  egocentric views.   (d)
    Illuminations.  Like  the rubber duck  in these examples, the  appearance of
    objects  can vary  dramatically with  illumination.  (e)  Ambiguities.  Many
    manufactured objects  have symmetric or  almost symmetric shapes,  which may
    make  pose  estimation   ambiguous  (object  and  images   from  the  T-Less
    dataset~\protect\cite{Hodan16}).}
\end{figure}

There are  many challenges that need  to be tackled in  practice when estimating
the 3D  poses of objects  and hands  from images.  We  list below some  of these
challenges, illustrated in Figure~\ref{fig:challenges}.

\paragraph{High degrees of freedom. } As mentioned above, the pose of a rigid
object can be represented with 6 scalar values, which is already a large number
of degrees of freedom to estimate. The 3D pose of a human hand lives in an even
much higher dimensional space, as it is often represented with about 32 degrees
of freedom. The risk of making an error when estimating the pose increases with
this number of degrees of freedom.

\paragraph{Occlusions. } Occlusions of the target objects or hands, even
partial,  often disturb  pose  estimation  algorithms. It  can  be difficult  to
identify which parts are visible and  which parts are occluded, and the presence
of occlusions may  make pose estimation completely fail, or  be very inaccurate.
Occlusions often  happen in practice.  For  example, when a hand  manipulates an
object, both the hand and the object are usually partially occluded.

\paragraph{Cluttered background. } Objects in the background can act as
distractors, especially if they look like target objects, or have similar parts.

\paragraph{Changing illumination conditions.} In practice, illumination cannot
be controlled, and  will change the appearance of the  objects, not only because
the lighting is  different but also because  shadows can be cast  on them.  This
requires  robust algorithms.   Moreover, cameras  may  struggle to  keep a  good
balance  between bright  glares and  dark areas,  and using  high dynamic  range
cameras becomes  appealing under such  conditions.  Moreover, sunlight  may make
depth cameras fail as discussed above.

\paragraph{Material and Textures.} Early approaches to pose estimation relied on
the presence  of texture  or pattern  on the  objects' surfaces,  because stable
features  can be  detected  and  matched relatively  easily  and efficiently  on
patterns. However, in practice, many objects lack textures, making such approach
fail. Non-Lambertian  surfaces such as  metal and  glass make the  appearance of
objects change with the objects'  poses because of specularities and reflections
appearing  on the  objects' surfaces.   Transparent objects  are also  of course
problematic, as they do not appear clearly in images.

\paragraph{Ambiguities.} Many manufactured objects are symmetrical or almost
symmetrical, or  have repetitive patterns.  This  generates possible ambiguities
when estimating the poses of these objects that need to be handled explicitly.

\paragraph{3D Model Requirement. } Most of the existing algorithms for pose
estimation assume the  knowledge of some 3D models of  the target objects.  Such
3D models provide  very useful geometric constraints, which can  be exploited to
estimate the objects' 3D poses. However,  building such 3D models takes time and
expertise, and  in practice,  a 3D  model is not  readily available.   Hand pose
estimation suffers much  less from this problem, since the  3D geometry of hands
remains very  similar from one person  to another.  Some recent  works have also
considered  the  estimation  of  the   objects'  geometry  together  with  their
poses~\cite{Grabner18}.  These  works   are  still  limited  to   a  few  object
categories,  such  as chairs  or  cars,  but  are  very interesting  for  future
developments.

%% file: early_approaches.tex
3D pose  estimation from  images has  a long history  in computer  vision. Early
methods were  based on  simple image  features, such as  edges or  corners. Most
importantly, they strongly relied on some  prior knowledge about the 3D pose, to
guide the pose estimation in the  high-dimensional space of possible poses. This
prior may  come from the  poses estimated for the  previous images, or  could be
provided by a human operator.

\paragraph{3D tracking methods. }
For example,  pioneer works such  as \cite{Harris90} and  \cite{Lowe91} describe
the object  of interest as a  set of 3D  geometric primitives such as  lines and
conics, which were matched with contours in the image to find a 3D pose estimate
by solving an  optimization problem to find  the 3D pose that  reprojects the 3D
geometric  primitives to  the matched  image  contours.  The  entire process  is
computationally light,  and careful  implementations were  able to  achieve high
frame rates with computers that would appear primitive to us.

Unfortunately, edge-based approaches are  quite unreliable in practice. Matching
the  reprojection of  the  3D primitives  with image  contours  is difficult  to
achieve:  3D primitives  do not  necessarily  appear as  strong image  contours,
except for carefully chosen objects. As a result, these primitives are likely to
be  matched with  the  wrong image  contours, especially  in  case of  cluttered
background.  Incorrect  matches will also  occur in case of  partial occlusions.
Introducing  robust estimators  into the  optimization problem~\cite{Drummond02}
helps, but  it appears that  it is  impossible in general  to be robust  to such
mismatches in edge-based pose estimation: Most  of the time, a match between two
contours provides  only limited constraints on  the pose parameters, as  the two
contours can ``slide'' along each other and still be a good match. Contours thus
do not provide reliable constraints for pose estimation.

Relying  on feature  points~\cite{Harris88}  instead of  contours provides  more
constraints as point-to-point  matching does not have  the ``sliding ambiguity''
of contour matching. Feature point matching has been used for 3D object tracking
with some  success, for example  in \cite{Vacchetti04}.  However,  such approach
assumes the  presence of  feature points  that can be  detected on  the object's
surface, which is not true for all the objects.

\paragraph{The importance of detection methods. }
The two  approaches described above  assume that  prior knowledge on  the object
pose is  available to  guide the  matching process  between contours  or points.
Typically, the  object pose estimated at  time $t$ is exploited  to estimate the
pose at time $t+1$. In practice, this makes such approaches fragile, because the
poses at time $t$  and $t+1$ can be very different if the  object moves fast, or
because the pose estimated at time $t$ can be wrong.

Being able  to estimate the  3D pose  of an object  without relying too  much on
prior knowledge is  therefore very important in practice.  As  we will see, this
does not mean that  methods based on strong prior knowledge on  the pose are not
useful.   In  fact, they  tend  to  be much  faster  and/or  more accurate  than
``detection methods'',  which are more  robust.  A  natural solution is  thus to
combine  both, and  this is  still true  in the  modern era  of object  and pose
estimation.

One early popular method for object  pose estimation without pose prior was also
based on  feature points, often referred  to as keypoints in  this context. This
requires the ability to match keypoints  between an input image, and a reference
image of the target  object, which is captured offline and in  which the 3D pose
and the 3D  shape of the object  is known. By using geometry  constraints, it is
then possible  to estimate the object's  3D pose.  However, wide  baseline point
matching is  much more difficult than  short baseline matching used  by tracking
methods.    SIFT   keypoints   and   descriptors~\cite{Lowe01,Lowe04}   were   a
breakthrough that made many computer  vision applications possible, including 3D
pose   estimation.    They   were   followed  by   faster   methods,   including
SURF~\cite{Bay08}  and ORB~\cite{Rublee11}.

As  for   tracking  methods   based  on  feature   points,  the   limitation  of
keypoint-based detection  and pose estimation is  that it is limited  to objects
exhibiting enough keypoints, which is not the case in general. This approach was
still very successful for Augmented Reality  in magazines for example, where the
``object''  is  an   image  printed  on  paper---so  it  has   a  simple  planar
geometry---and  selected  to guarantee  that  the  approach  will work  well  by
exhibiting enough keypoints~\cite{Kim10}.

To  be able  to  detect and  estimate  the 3D  pose of  objects  with almost  no
keypoints,  sometimes  referred  to  as  ``texture-less''  objects,  some  works
attempted to use ``templates'': The  templates of \cite{Hinterstoisser12} aim at
capturing the  possible appearances of  the object  by discretizing the  3D pose
space, and representing  the object's appearance for each discretized  pose by a
template  covering  the  full object,  in  a  way  that  is robust  to  lighting
variations. Then, by scanning the input images looking for templates, the target
object can be detected  in 2D and its pose estimated based  on the template that
matches  best  the  object  appearance.  However,  such  approach  requires  the
creation of many templates, and is poorly robust to occlusions.

\paragraph{Conclusion. }
We  focused in  this section  on object  pose estimation  rather than  hand pose
estimation, however  the conclusion  would be the  same.  Early  approaches were
based on handcrafted  methods to extract features from images,  with the goal of
estimating the 3D pose from these  features. This is however very challenging to
do,  and  almost  doomed to  fail  in  the  general  case. Since  then,  Machine
Learning-based methods  have been shown  to be more  adapted, even if  they also
come with their drawbacks, and will be discussed in the rest of this chapter.

%% file: ml_dl.tex
Fundamentally, 3D pose estimation  of objects or hands can be  seen as a mapping
from an image to  a representation of the pose. The input  space of this mapping
is thus the  space of possible images,  which is an incredibly  large space: For
example, a  RGB VGA image is  made of almost  1 million values of  pixel values.
Not many fields  deal with 1 million  dimension data!  The output  space is much
smaller, since it is made  of 6 values for the pose of a  rigid object, or a few
tens for the pose of a hand.  The  natures of the input and output spaces of the
mapping sought in pose estimation are therefore very different, which makes this
mapping very  complex.  From this  point of view, we  can understand that  it is
very difficult to hope for a pure ``algorithmic'' approach to code this mapping.

This is why  Machine Learning techniques, which use data  to improve algorithms,
have  become  successful  for  pose estimation  problems,  and  computer  vision
problems in general.  Because they  are data-driven, they can find automatically
an  appropriate mapping,  by  contrast with  previous  approaches that  required
hardcoding mostly based on intuition, which can be correct or wrong.

Many  Machine Learning  methods exist,  and Random  Forests, also  called Random
Decision  Forests or  Randomized  Trees, were  an early  popular  method in  the
context  of 3D  pose estimation~\cite{Lepetit05,Shotton12,Brachmann14}.   Random
Forests can be  efficiently applied to image  patches and discrete~(multi-class)
and continuous~(regression) problems, which makes  them flexible and suitable to
fast, possibly real-time, applications.

\paragraph{Deep Learning. }
For multiple reasons, Deep Learning, a Machine Learning technique, took over the
other methods almost  entirely during the last years in  many scientific fields,
including 3D  pose estimation.  It  is very flexible, and  in fact, it  has been
known  for a  long  time that  any  continuous mapping  can  be approximated  by
two-layer networks, as finely  as wanted~\cite{Hornik89,Pinkus99}.  In practice,
networks  with more  than two  layers  tend to  generalize better,  and to  need
dramatically less parameters than two-layer networks~\cite{Eldan16}, which makes
them a tool of choice for computer vision problems.

Many  resources can  now  be easily  found  to learn  the  fundamentals of  Deep
Learning, for example \cite{GoodFellow16}.  To stay  brief, we can say here that
a Deep Network can be defined as a composition of functions~(``layers'').  These
functions may depend  on parameters, which need to be  estimated for the network
to perform  well.  Almost any function  can be used as  layer, as long as  it is
useful to solve  the problem at hand, and if  it is \emph{differentiable}.  This
differentiable  property  is  indeed  required  to  find  good  values  for  the
parameters by solving an optimization problem.

\begin{figure}
  \begin{center}
    \includegraphics[width=0.8\linewidth]{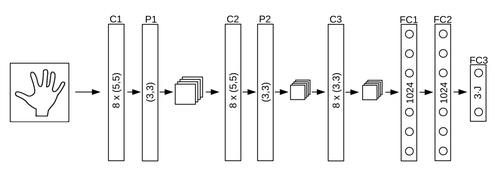} 
  \end{center}
  \caption{\label{fig:basic_dn}  A basic  Deep Network  architecture applied  to
    hand  pose  prediction.   \textsf{C1},   \textsf{C2},  and  \textsf{C3}  are
    convolutional layers,  \textsf{P1} and  \textsf{P2} are pooling  layers, and
    \textsf{FC1}, \textsf{FC2}, and \textsf{FC3} are fully connected layers. The
    numbers in  each bar  indicates either:  The number and  size of  the linear
    filters in case of the convolutional  layers, the size of the pooling region
    for the  pooling layers, and the  size of the  output vector in case  of the
    fully connected layers.}
\end{figure}

\paragraph{Deep Network Training.}
For example, if we want to make a network $F$ predict the pose of a hand visible
in  an  image, one  way  to  find good  values  $\hat{\Theta}$  for the  network
parameters is to solve the following optimization problem:
\begin{equation}
  \begin{array}{l}
    \hat{\Theta} = \arg \min_\Theta \calL(\Theta) \>  \text{  with}\\[0.2cm]
    \calL(\Theta) = \frac{1}{N} \sum\limits_{i=1}^N \|F(I_i;\Theta) - \be_i\|^2 \> ,
  \end{array}
  \label{eq:opt}
\end{equation}
where $\{(I_1, \be_1),  .., (I_N, \be_N)\}$ is a  \emph{training set} containing
pairs  made of  images  $I_i$ and  the corresponding  poses  $\be_i$, where  the
$\be_i$ are vectors that contain, for example, the 3D locations of the joints of
the hand as  was done in~\cite{Oberweger17}.  Function $\calL$ is  called a loss
function.  Optimizing the network parameters  $\Theta$ is called training.  Many
optimization algorithms  and tricks  have been proposed  to solve  problems like
Eq.~\eqref{eq:opt}, and many software libraries exist to make the implementation
of the network creation and its training an easy task.

Because  optimization algorithms  for  network training  are  based on  gradient
descent, any differentiable loss function can  be used in principle, which makes
Deep Learning  a very  flexible approach,  as it  can be  adapted easily  to the
problem at hand.

\paragraph{Supervised Training and the Requirement for Annotated Training Sets. }
Training a network by using a loss  function like the one in Eq.~\ref{eq:opt} is
called supervised training,  because we assume the availability  of an annotated
dataset of  images.  Supervised training tends  to perform very well,  and it is
used very  often in practical  applications.

This  is however  probably  the main  drawback of  Deep  Learning-based 3D  pose
estimation.  While early methods relied only on a 3D model, and for some of them
only on a small number of images  of the object, modern approaches based on Deep
Learning require a large dataset of images of the target objects, annotated with
the ground truth 3D poses

Such datasets are already available~(see Section~\ref{sec:datasets}), but
they are useful mostly for evaluation and comparison purposes.  Applying current
methods to  new objects requires creating  a dataset for these  new objects, and
this is a cumbersome task.

It is  also possible to  use synthetic data  for training, by  generating images
using  Computer Graphics  techniques. This  is used  in many  works, often  with
special to take into account the  differences between real and synthetic images,
as we will see below.

%% file: object_datasets.tex
\begin{figure}
  \begin{center}
    \begin{tabular}{cccc}
      \includegraphics[height=0.16\linewidth]{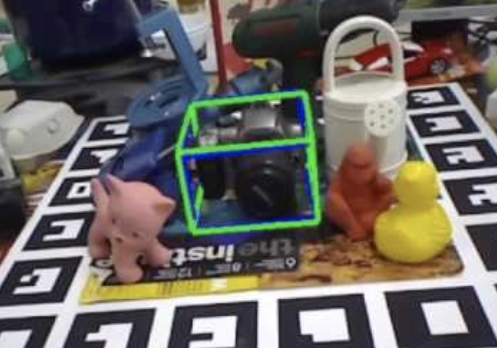}&
      \includegraphics[height=0.16\linewidth]{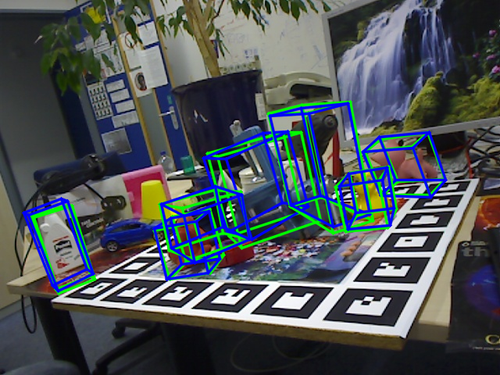}&
      \includegraphics[height=0.16\linewidth]{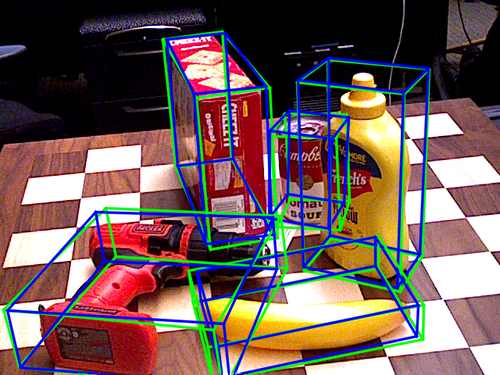}&
      \includegraphics[height=0.16\linewidth]{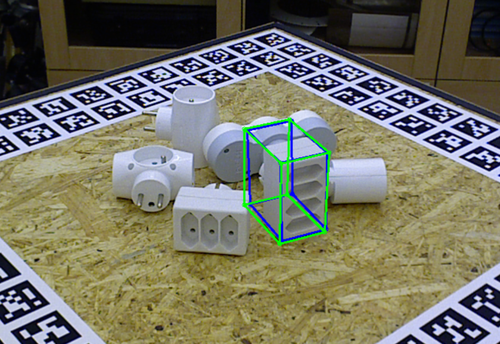}\\
      (a) & (b) & (c) & (d)\\
\end{tabular}
\end{center}
\caption{\label{fig:object_datasets} Images from popular  datasets for 3D object
  pose  estimation.  (a)  LineMOD;  (b) Occluded  LineMOD;  (c)  YCB-Video;  (d)
  T-Less.}
\end{figure}

\paragraph{LineMOD Dataset.} The LineMOD dataset~\cite{Hinterstoisser12}
predates most machine learning approaches and as  such, it is not divided into a
training and a  test sets. It is made  of 15 small objects, such as  a camera, a
lamp, and a cup.   For each object, it offers a set of  1200 RGB-D images of the
object  surrounded by  clutter.  The  other objects  are  often  visible in  the
clutter, but  only the 3D pose  of the target  object is provided for  each set.
The 3D models of the objects are also provided.

\paragraph{Occluded LineMOD Dataset. } The Occluded LineMOD dataset
was created by  the authors of~\cite{Brachmann14}  from LineMOD by annotating the  3D poses of
the objects belonging  to the dataset but originally not  annotated because they
were considered  as part of  the clutter. This results  into a sequence  of 1215
frames, each frame labeled with the 3D  poses of eight objects in totla, as well
as the  objects' masks.  The  objects show  severe occlusions, which  makes pose
estimation challenging.

\paragraph{YCB-Video Dataset. } This dataset~\cite{Xiang18b} consists of 92 video
sequences,  where  12 sequences  are  used  for  testing  and the  remaining  80
sequences for  training.  In  addition, the  dataset contains  80k synthetically
rendered images, which  can be used for  training as well. There  are 21 ``daily
life'' objects  in the dataset, from  cereal boxes to scissors  or plates. These
objects were selected from the  YCB dataset~\cite{Calli17} and are available for
purchase.  The dataset  is captured with two different RGB-D  sensors.  The test
images are challenging due to the presence of significant image noise, different
illumination levels, and large occlusions.  Each  image is annotated with the 3D
object poses, as well as the objects' masks.

\paragraph{T-Less Dataset. } The T-Less dataset~\cite{Hodan16} is made from 30
``industry-relevant'' objects.   These objects have no  discriminative color nor
texture.  They  present different types  of symmetries and  similarities between
them, making  pose estimation often  almost ambiguous. The images  were captured
using three synchronized sensors: two  RGB-D cameras, one structured-light based
and  one time-of-flight  based, and  one  high-resolution RGB  camera. The  test
images~(10K from each  sensor) are from 20 scenes  with increasing difficulties,
with  partial occlusions  and contacts  between objects.   This dataset  remains
extremely challenging.

%% file: hand_datasets.tex
\paragraph{Early datasets.}
The NYU  dataset~\cite{Tompson14} contains over  72k training and 8k  test RGB-D
images data, captured  from three different viewpoints  using a structured-light
camera.  The images were annotated with 3D joint locations with a semi-automated
approach, by using a standard  3D hand tracking algorithm reinitialized manually
in case of failure.  The  ICVL dataset~\cite{Tang14} contains over 180k training
depth  frames showing  various  hand poses,  and two  test  sequences with  each
approximately 700 frames,  all captured with a time-of-flight  camera. The depth
images  have a  high quality  with  hardly any  missing depth  values and  sharp
outlines with  little noise.  Unfortunately,  the hand pose variability  of this
dataset  is limited  compared  to  other datasets,  and  annotations are  rather
inaccurate~\cite{Supancic15}.  The MSRA  dataset~\cite{Sun15} contains about 76k
depth  frames,  captured  using  a time-of-flight  camera  from  nine  different
subjects.

\paragraph{BigHands Dataset. }
The  BigHands dataset~\cite{Yuan17}  contains an  impressive 2.2  millions RGB-D
images captured with  a structured-light camera.  The  dataset was automatically
annotated  by using  six 6D  electromagnetic sensors  and inverse  kinematics to
provide  21  3D joint  locations  per  frame.  A  significant  part  is made  of
egocentric views. The  depth images have a high quality  with hardly any missing
depth values,  and sharp outlines with  little noise.  The labels  are sometimes
inaccurate because of  the annotation process, but the dataset  has a large hand
pose variability, from ten users.

\paragraph{CMU Panoptic Hand Dataset. }
The CMU Panoptic  hand dataset~\cite{Simon17} is made of 15k  real and synthetic
RGB images, from a third-person point of view.  The real images were recorded in
the CMU's  Panoptic studio and  annotated by the  method proposed in  the paper,
based on multiple views.  The annotations are only in 2D but can still be useful
for multi-view pose estimation.

%% file: hand_object_datasets.tex
\begin{figure}
  \begin{center}
    \begin{tabular}{ccc}
      \includegraphics[width=0.3\linewidth]{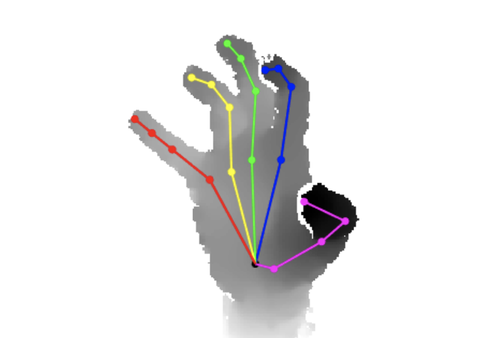}&
      \includegraphics[width=0.2\linewidth]{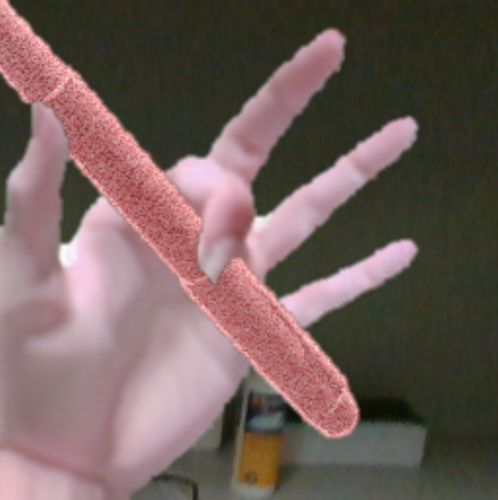}&
      \includegraphics[width=0.3\linewidth]{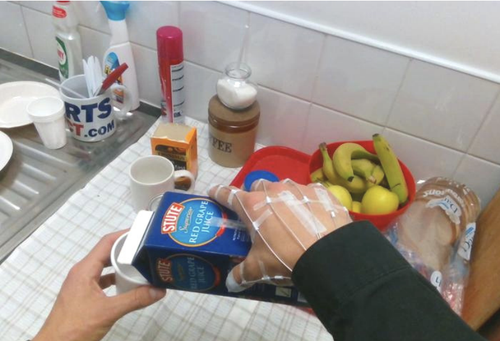}\\
      (a) & (b) & (c)\\
      \includegraphics[width=0.3\linewidth]{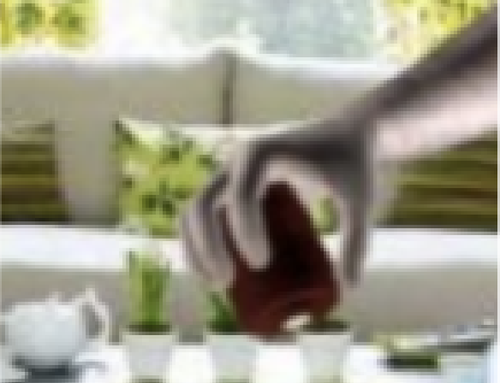}&
      \includegraphics[width=0.3\linewidth]{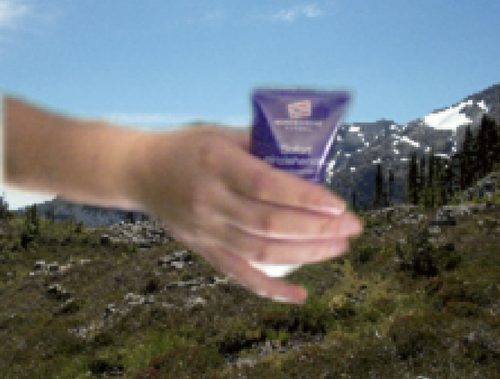}&
      \includegraphics[width=0.3\linewidth]{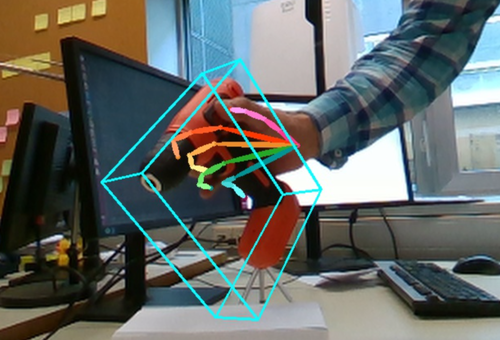}\\
      (d) & (e) & (f)\\
    \end{tabular}
  \end{center}
  \caption{\label{fig:hand_datasets} Images  from popular  datasets for  3D hand
    (a)  and  hand+object (b-f)  pose  estimation.   (a) BigHand;  (b)
    GANerated hand dataset; (c) First-Person Hand Action dataset; (d) Obman
    dataset; (e) FreiHAND dataset; (f) HO-3D dataset.}
\end{figure}

\paragraph{GANerated Hand Dataset. }
The GANerated  hand dataset~\cite{Mueller2018} is  a large dataset made  of 330K
synthetic images  of hands, sometimes  holding an object,  in front of  a random
background.  The images are annotated with the  3D poses of the hand. The images
were made more realistic by extending CycleGAN~\cite{Zhu17}.

\paragraph{First-Person Hand Action dataset. }
The First-Person Hand Action dataset~\cite{Garcia2018first} provides a
dataset of hand  and object interactions with 3D  annotations for both
hand joints and  object pose.  They used a motion  capture system made
of  magnetic sensors  attached to  the  user's hand  and to  the
object in  order to  obtain hand  3D pose  annotations in  RGB-D video
sequences.  Unfortunately, this changes the  appearance of the hand in
the  color images  as  the sensors  and the  tape  attaching them  are
visible,  but the  dataset proposes  a  large number  of frames  under
various  conditions (more  than 100K  egocentric views  of 6  subjects
doing 45 different types of daily-life activities).

\paragraph{ObMan Dataset. }
Very recently,  \cite{Hasson2019learning} introduced  ObMan, a large  dataset of
images of hands  grasping objects. The images are synthetic,  but the grasps are
generated using an algorithm from robotics  and the grasps still look realistic.
The dataset provides the  3D poses and shapes of the hand as  well as the object
shapes.

\paragraph{FreiHand Dataset. }
\cite{zimmermann2019freihand} proposed a multi-view RGB dataset, FreiHAND, which
includes hand-object  interactions and provides the  3D poses and shapes  of the
hand.  It relies on a green-screen background  environment so that it is easy to
change the background for training purposes.

\paragraph{HO-3D Dataset. }
\cite{Hampali20}  proposed a  method to  automatically annotate  video sequences
captured with  one or  more RGB-D  cameras with  the object  and hand  poses and
shapes. This  results in  a dataset made  of 75,000 real  RGB-D images,  from 10
different  objects and  10  different  users.  The  objects  come  from the  YCB
dataset~(see  Section~\ref{sub:object_datasets}). The  backgrounds are  complex,
and  the mutual  occlusions are  often large,  which makes  the pose  estimation
realistic but very challenging.

%% file: metrics.tex
Metrics are important to evaluate and  compare methods.  Many metrics exist, and
we describe here only the main  ones for object pose estimation.  Discussions on
metrics for 3D object pose estimation can be found in \cite{Hodan16} and
\cite{Bregier18}.

\paragraph{ADD, ADI, ADD-S, and the 6D Pose metrics.}
The ADD  metric~\cite{Hinterstoisser12b} calculates  the average distance  in 3D
between the model points, after applying the ground truth pose and the predicted
pose. This can be formalized as:
\begin{equation}
  \ADD = \frac{1}{|\calV|}\sum_{\bM \in \calV}{\| \RigidMotion(\bM; \hat{\pose}) - \RigidMotion(\bM; \bar{\pose}) \|_2} \> ,
\end{equation}
where $\calV$ is  the set of the object's vertices,  $\hat{\pose}$ the estimated
pose and $\bar{\pose}$ the ground truth pose, and $\RigidMotion(\bM; \pose)$ the
rigid  transformation in  $\pose$ applied  to 3D  point $\bM$.   In the  6D Pose
metric, a pose is  considered when the ADD metric is less  than $10\%$ of the
object's diameter.

For the objects with ambiguous poses due to symmetries, \cite{Hinterstoisser12b}
replaces the ADD metric by the ADI  metric, also referred to as the ADD-S metric
in~\cite{Xiang2018}, computed as follows:
\begin{equation}
  \text{ADD-S} =  \frac{1}{|\calV|}\sum_{\bM_1 \in \calV}{\min_{\bM_2 \in \calV}{\|
      \RigidMotion(\bM_1; \hat{\pose}) - \RigidMotion(\bM_2; \bar{\pose})
      \|_2}} \> ,
  \label{eq:ADI}
\end{equation}
which averages  the distances from points  after applying the predicted  pose to
the \emph{closest}  points under the  ground truth  pose. The advantage  of this
metric is  that it is indeed  equal to zero when  the pose is retrieved  up to a
symmetry, even if it does not exploit the symmetries of the object.

%% file: object_pose.tex
Over the past years, many authors realise  that Deep Learning is a powerful tool
for 3D object  pose estimation from images.  We discuss  here the development of
Deep Learning applied to 3D object  pose estimation over time.  This development
was  and is  still  extremely  fast, with  improving  accuracy, robustness,  and
computation  times.   We  present  this development  through  several  milestone
methods, but much more methods could also be included here.

\begin{figure}
  \begin{center}
    \includegraphics[width=0.8\linewidth]{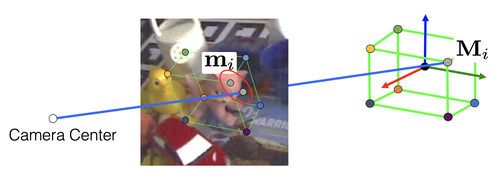}
  \end{center}
  \caption{\label{fig:bb8} Some 3D object pose estimation methods predict the pose
    by first  predicting the 2D reprojections  $\bm_i$ of some 3D  points $\bM_i$,
    and  then   computing  the  3D   rotation  and  translation  from   the  2D-3D
    correspondences  between  the  $\bm_i$  and   $\bM_i$  points  using  a  P$n$P
    algorithm.}
\end{figure}

\subsection{BB8}
\label{sec:bb8}

One of the first Deep Learning methods for 3D object pose estimation is probably
BB8~\cite{Rad17}.  As it is the first method  we describe, we will present it in
some details.

This method proceeds  in three steps.
It first  detect the target objects  in 2D using coarse  object segmentation. It
then  applies  a  Deep  Network  on  each  image  window  centered  on  detected
objects. Instead of predicting  the 3D pose of the detected  objects in the form
of a  3D translation and  a 3D  rotation, it predict  the 2D projections  of the
corners of the object's  bounding box, and compute the 3D  pose from these 2D-3D
correspondences  with a  P$n$P algorithm~\cite{Gao03}---hence  the name  for the
method,  from   the  8  corners  of   the  bounding  box,  as   illustrating  in
Figure~\ref{fig:bb8}.   Compared to  the  direct prediction  of  the pose,  this
avoids the  need for a  meta-parameter to  balance the translation  and rotation
terms.  It also  tends to make network optimization  easier. This representation
was used in some later works.

Since it is the  first Deep Learning method for pose  estimation we describe, we
will detail  the loss function  (see Section~\ref{sec:ml_dl}) used to  train the
network.   This loss  function  $\calL$  is a  least-squares  error between  the
predicted 2D points and the expected ones, and can be written as:
\begin{equation}
  \calL(\Theta) = \frac{1}{8} \sum_{(W, \pose) \in \calT} \> \sum_i \| \Proj_\pose(\bM_i) -  F(W;\Theta)_i\|^2 \> ,
  \label{eq:pose_training}
\end{equation}
where $F$ denotes the trained network and $\Theta$ its parameters.  $\calT$ is a
training set made of  image windows $W$ containing a target  object under a pose
$\pose$. The $\bM_i$ are  the 3D coordinates of the corners  of the bounding box
of for  this object, in  the object coordinate system.   $\Proj_{\be, \bt}(\bM)$
projects the  3D point $\bM$  on the  image from the  pose defined by  $\be$ and
$\bt$.   $F(W; \Theta)_i$  returns  the  two components  of  the  output of  $F$
corresponding to the predicted 2D coordinates of the $i$-th corner.

\paragraph{The problem of symmetrical and ``almost symmetrical'' objects. }
Predicting the  3D pose  of objects  a standard  least-squares problem,  using a
standard representation of the pose or  point reprojections as in BB8, yields to
large  errors  on symmetrical  objects,  such  as  many  objects in  the  T-Less
dataset~(Figure~\ref{fig:object_datasets}(d)).    This   dataset  is   made   of
manufactured objects that are not only similar  to each other, but also have one
axis of rotational symmetry. Some objects are not perfectly symmetrical but only
because of small details, like a screw.

The approach described above fails on these  objects because it tries to learn a
mapping  from  the image  space  to  the pose  space.   Since  two images  of  a
symmetrical  object under  two different  poses look  identical, the  image-pose
correspondence is  in fact a  one-to-many relationship.

For objects  that are  perfectly symmetrical,  \cite{Rad17} proposed  a solution
that we will  not describe here, as simpler solutions  have been proposed since.
Objects that are ``almost symmetrical'' can also disturb pose prediction, as the
mapping from the image to the 3D pose  is difficult to learn even though this is
a one-to-one mapping.  Most recent methods ignore the problem  and consider that
these objects are  actually perfectly symmetrical and consider  a pose recovered
up to the symmetries as correct. For such object, BB8 proposes to first consider
these  objects as  symmetrical,  and then  to train  a  classifier~(also a  Deep
Network) to predict which pose is actually  the correct one. For example, for an
object with a rotational symmetry of  $180\degree$, there are two possible poses
in general, and the classifier has to  decide between 2 classes.  This is a much
simpler  problem than  predicting 6  degrees-of-freedom  for the  pose, and  the
classifier can focus on the small details that break the symmetry.

\begin{figure}
  \begin{center}
      \includegraphics[width=0.5\linewidth]{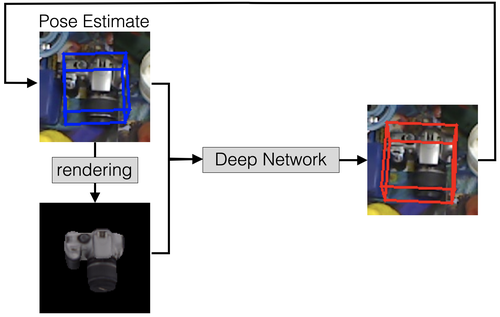}
  \end{center}
  \caption{\label{fig:bb8_refinement}  Pose refinement  in BB8.   Given a  first
    pose  estimate, shown  by  the  blue bounding  box,  BB8  generates a  color
    rendering of the object.  A network is  trained to predict an update for the
    object  pose given  the input  image  and this  rendering, to  get a  better
    estimate shown by the red bounding box.  This process can be iterated.}
\end{figure}

\paragraph{Refinement step. }
The method  described above provides  an estimate for the  3D pose of  an object
using only feedforward computation  by a network from the input  image to the 3D
pose. It is relatively natural to aim at refining this estimate, which is a step
also present  in many other  methods. In BB8, this  is performed using  a method
similar to the  one proposed in \cite{Oberweger15b} for hand  detection in depth
images.  A network is trained to improve the prediction of the 2D projections by
comparing the  input image and  a rendering of the  object for the  initial pose
estimate, as illustrated in Figure~\ref{fig:bb8_refinement}. Such refinement can
 be iterated multiple times to improve the pose estimate.

One may  wonder why the refinement  step, in BB8  but also in more  recent works
such  as DeepIM~\cite{Xiang2018}  or  DPOD~\cite{Zakharov19},  can improve  pose
accuracy while it is (apparently) trained with the same data as the initial pose
prediction. This  can be understood by  looking more closely at  how the network
predicting the update is trained. The input  part of one training sample is made
of a regular  input image, plus an rendered  image for a pose close  to the pose
for input image,  and the output part  is the difference between  the two poses.
In practice, the pose  for the rendered image is taken as  the ground truth pose
for the real image  plus some random noise.  In other words,  from one sample of
the  original  dataset  trained  to  predict the  network  providing  the  first
estimate, it is possible to generate  a virtually infinite number of samples for
training the  refinement network,  by simply  adding noise  to the  ground truth
pose.  The  refinement network is thus  trained with much more  samples than the
first network.

\paragraph{Data augmentation. }
Data augmentation,  \emph{i.e.} generating  additional training images  from the
available ones,  is often critical  in Deep Learning.  In  the case of  BB8, the
objects' silhouettes are extracted from  the original training images, which can
be done using the ground truth poses and the objects' 3D models.  The background
is replaced by a patch extracted from  a randomly picked image from the ImageNet
dataset~\cite{Russakovsky15}.   Note  that  this procedure  removes  context  by
randomly replacing  the surroundings of  the objects.  Some information  is thus
lost, as context could be useful for pose estimation.

\subsection{SSD-6D}%

\begin{figure}
  \begin{center}
    \begin{tabular}{cc}
    \includegraphics[height=0.23\linewidth]{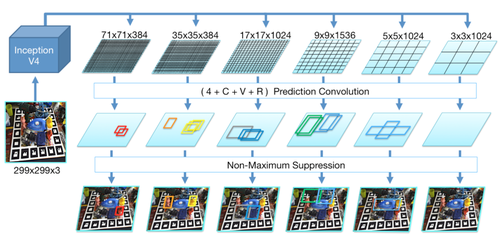} &
\includegraphics[height=0.23\linewidth]{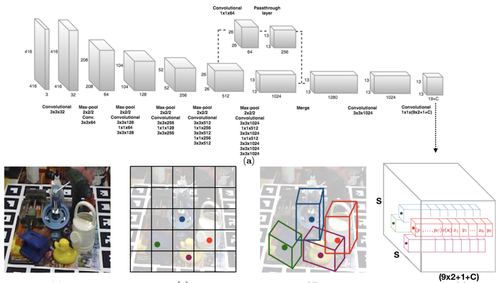} \\
(a) & (b)\\
\end{tabular}
  \end{center}
  \caption{\label{fig:ssd6d_yolo6d} (a) The architecture of  SSD, extended by SSD-6D to
    predict 3D poses.  For each bounding box corresponding to a detected object,
    a  discretized   3D  pose   is  also  predicted   by  the   network.   Image
    from~\protect\cite{Kehl17}.   (b)   The  architecture  of   YOLO-6D.   Image
    from~\protect\cite{Tekin18}.}
\end{figure}

BB8 relies on two separate networks to first detect the target objects in 2D and
then  predict their  3D poses,  plus  a third  one if  refinement is  performed.
Instead,  as  shown  in  Figure~\ref{fig:ssd6d_yolo6d}(a),  SSD-6D~\cite{Kehl17}
extends a deep architecture~(the SSD architecture~\cite{Liu16}) developed for 2D
object detection to 3D pose estimation~(referred in the paper as 6D estimation).
SSD had already been extended to pose estimation in~\cite{Poirson16}, but SS6-6D
performs full 3D pose prediction.

This prediction is done by first  discretizing the pose space. Each discretized
pose is considered as a class, to turn the pose prediction into a classification
problem  rather than  a regression  one, as  this was  performing better  in the
authors' experience. This  discretization was done on the  3D pose decomposition
into  direction of  view  over  a half-sphere  and  in-plane  rotation.  The  3D
translation can be computed from the  2D bounding box.  To deal with symmetrical
objects, views on the half-sphere  corresponding to identical appearance for the
object are merged into the same class.

A single network is therefore trained to perform both 2D object detection and 3D
pose  prediction,  using a  single  loss  function made  of  a  weighted sum  of
different terms. The weights are hyperparameters and have to be tuned, which can
be difficult. However, having a single  network is an elegant solution, and more
importantly for practical  applications, this allows to  save computation times:
Image feature  extraction, the slowest  part of  the network, is  performed only
once  even if  it is  used to  predict the  2D bounding  boxes and  the 3D  pose
estimation. 

A refinement procedure is then run on  each object detection to improve the pose
estimate predicted by the classifier.  SSD-6D  relies on a method inspired by an
early approach  to 3D  object tracking  based on  edges~\cite{Drummond02}.  Data
augmentation was done by rendering synthetic views of the objects using their 3D
models over images from COCO~\cite{Lin14a}, which  helps but only to some extend
as the  differences between the real  and synthetic images (the  ``domain gap'')
remain high.

\subsection{YOLO-6D}

The  method  proposed  in~\cite{Tekin18},  sometimes  referred  as  YOLO-6D,  is
relatively similar to  SSD-6D, but makes different choices that  makes it faster
and more accurate.   As shown in Figure~\ref{fig:ssd6d_yolo6d}(b),  it relies on
the  YOLO architecture~\cite{Redmon16,Redmon17}  for  2D  object detection,  and
predicts the 3D object  poses in a form similar to the one  of BB8.  The authors
report that training the network using this pose representation was much simpler
than  when using  quaternions to  represent the  3D rotation.   As YOLO  is much
faster and as they do not discretize the pose space, \cite{Tekin18} also reports
much  better performance  than SSD-6D  in terms  of both  computation times  and
accuracy.

\subsection{PoseCNN}

\begin{figure}
\begin{center}
\includegraphics[width=0.8\linewidth]{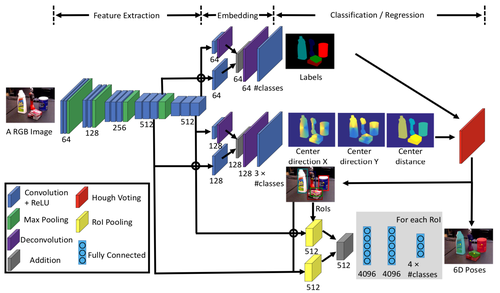}
\end{center}
\caption{\label{fig:posecnn}    The    architecture     of    PoseCNN.     Image
  from~\protect\cite{Xiang2018}.}
\end{figure}

PoseCNN is  a method  proposed in  \cite{Xiang2018}. It  relies on  a relatively
complex architecture,  based on the  idea of  decoupling the 3D  pose estimation
task  into  different  sub-tasks.   As shown  in  Figure~\ref{fig:posecnn},  this
architecture predicts for each pixel in the  input image 1) the object label, 2)
a unit vector towards  the 2D object center, and 3) the  3D distance between the
object center and the  camera center, plus 4) 2D bounding  boxes for the objects
and 5) a  3D rotation for each bounding  box in the form of  a quaternion.  From
1), 2),  and 3), it is  possible to compute  the 3D translation vector  for each
visible object,  which is combined  with the 3D rotation  to obtain the  full 3D
pose for each visible object.

Maybe the  most interesting contribution  of PoseCNN  is the loss  function.  It
uses the ADD  metric as the loss~(see Section~\ref{sec:metrics}),  and even more
interestingly, the paper shows that the  ADD-S metric, used to evaluate the pose
accuracy for symmetrical objects~(see Section~\ref{sec:object_datasets}), can be
used as a loss function to deal  with symmetrical objects.  The ADI metric makes
use  of  the $\min$  operator,  which  makes  it  maybe an  unconventional  loss
function, but  it is still  differentiable and can be  used to train  a network.
This results in an elegant and efficient way of handling object symmetries.

Another contribution  of \cite{Xiang2018} is  the introduction of  the YCB-Video
dataset~(see Section~\ref{sub:object_datasets}).

\subsection{DeepIM}

\begin{figure}
\begin{center}
\includegraphics[width=0.8\linewidth]{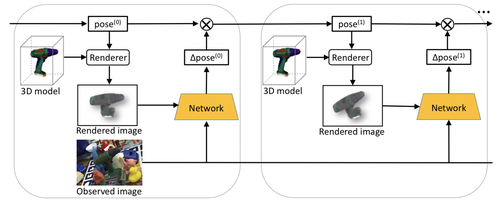}
\end{center}
\caption{\label{fig:deepim}   The   DeepIM~\protect\cite{Xiang2018}   refinement
  architecture.       It      is      similar       to      the      one      of
  BB8~(Figure~\ref{fig:bb8_refinement})  but   predicts  a  pose  update   in  a
  carefully chosen coordinate system for better generalization. }
\end{figure}

DeepIM~\cite{Li18} proposes a refinement step that resembles the one in BB8 (see
the discussion on refinement steps in Section~\ref{sec:bb8}), but
with several main differences.  As BB8, it trains a network to compare the input
image with  a rendering  of an object  that has already  been detected,  and for
which a  pose estimate is already  available. The network outputs  an update for
the 3D object pose that will improve  the pose estimate, and this process can be
iterated until convergence.  The first difference  with BB8 is that the rendered
image has a  high-resolution and the input  image is centered on  the object and
upscaled to the same resolution. This  allows a higher precision when predicting
the  pose update.  The loss  function also  includes terms  to make  the network
output the flow and the object mask, to introduce regularization.

The second difference is more fundamental,  as it introduces a coordinate system
well suited to define the rotation update that must be predicted by the network.
The authors first  remark that predicting the rotation in  the camera coordinate
system because this rotation would  also \emph{translate} the object.  They thus
set the center of rotation to the  object center. For the axes of the coordinate
system, the authors  remark that using those  of the object's 3D model  is not a
good option,  as they are  arbitrary and this would  force the network  to learn
them for  each object.   They therefore propose  to use the  axes of  the camera
coordinate  system,  which  makes  the  network  generalize  much  better.   The
translation update is predicted  as a 2D translation on the  image plane, plus a
delta along  the z axis  of the  camera in a  log-scale.  To learn  this update,
DeepIM  uses  the  same  loss  function as  \cite{Xiang2018}~(ADD,  or  ADD-S  for
symmetrical  objects).  This  approach is  even  shown to  generalize to  unseen
objects, but  this is admittedly demonstrated  in the paper only  on very simple
object renderings.

\subsection{Augmented Autoencoders}

\begin{figure}
\begin{center}
\includegraphics[width=0.8\linewidth]{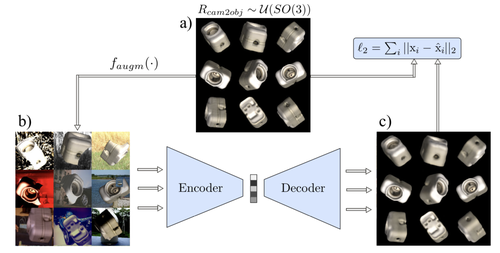}
\end{center}
\caption{\label{fig:augmented_autoencoders}  Autoencoders  trained to  learn  an
  image embedding. This embedding is robust  to nuisances such as background and
  illumination changes, and  the possible rotations for the object  in the input
  image   can   be    retrieved   based   on   this    embedding.   Image   from
  \protect\cite{Sundermeyer19}.}
\end{figure}

The  most  interesting aspect  of  the  Augmented Autoencoders  method  proposed
in~\cite{Sundermeyer19}  is  the way  it  deals  with symmetrical  objects.   It
proposes to first learn  an embedding that can be computed from  an image of the
object.   This embedding  should be  robust to  imaging artefacts~(illumination,
background, etc.)  and  depends only on the object's  appearance: The embeddings
for two images of the object under  ambiguous rotations should thus be the same.
At run-time,  the embedding for  an input  image can then  be mapped to  all the
possible rotations.  This is done efficiently by creating a codebook offline, by
sampling  views around  the target  objects, and  associating the  embeddings of
these views  to the  corresponding rotations. The  translation can  be recovered
from the object bounding box' 2D location and scale.

To  learn  to   compute  the  embeddings,  \cite{Sundermeyer19}   relies  on  an
autoencoder architecture, shown  in Figure~\ref{fig:augmented_autoencoders}.  An
Encoder network predicts an embedding from  an image of an object with different
image nuisances;  a Decoder network takes  the predicted embedding as  input and
should generate the image of the object  under the same rotation but without the
nuisances.  The encoder and decoder are  trained together on synthetic images of
the objects,  to which nuisances  are added.   The loss function  encourages the
composition of the two networks to output the original synthetic images, despite
using  the  images  with  nuisances  as  input.   Another  motivation  for  this
autoencoder architecture  is to learn to  be robust to nuisances,  even though a
more standard  approach outputting the 3D  pose would probably do  the same when
trained on the same images.

\subsection{Robustness to Partial Occlusions: \protect\cite{Oberweger18},
  \protect\cite{Hu19},  PVNet}

\begin{figure}
\begin{center}
\includegraphics[width=0.8\linewidth]{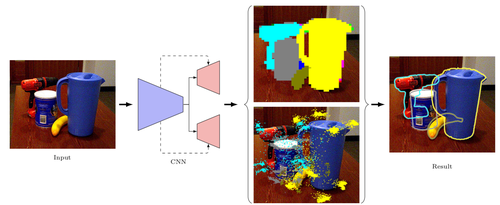}
\end{center}
\caption{\label{fig:segmentation_hu}  Being  robust  to  large  occlusions.  \protect\cite{Hu19}  makes
  predictions for  the 2D reprojections of  the objects' 3D bounding  boxes from
  many image  locations. Image locations  that are  not occluded will  result in
  good predictions,  the predictions made  from occluded image locations  can be
  filtered   out  using   a  robust   estimation  of   the  pose.   Image 
  from~\protect\cite{Hu19}.}
\end{figure}

\cite{Oberweger18},  \cite{Hu19}, and  PVNet~\cite{Peng19}  developed almost  in
parallel similar  methods that   provide accurate  3D poses even  under large
partial occlusions.  Indeed, \cite{Oberweger18} shows  that Deep Networks can be
robust to partial occlusions when predicting  a pose from a image containing the
target object when trained on examples with occlusions, but only to some extend.
To   become   more   robust   and  more   accurate   under   large   occlusions,
\cite{Oberweger18,Hu19,Peng19} proposed  to predict the  3D pose in the  form of
the 2D  reprojections of  some 3D points  as in BB8,  but by  combining multiple
predictions,  where each  prediction  is performed  from  different local  image
information.  As shown in Figure~\ref{fig:segmentation_hu}, the key idea is that
local image  information that is  not disturbed  by occlusions will  result into
good predictions; local  image information disturbed by  occlusions will predict
erroneous reprojections, but these can be filtered out with a robust estimation.

\cite{Oberweger18} predicts  the 2D  reprojections in the  form of  heatmaps, to
handle the ambiguities of mapping between image locations and the reprojections,
as many  image locations  can look  the same.   However, this  makes predictions
relatively slow.  Instead, \cite{Hu19} predicts a single 2D displacement between
the image  location and each  2D reprojection.   These results in  many possible
reprojections,  some noisy,  but  the correct  3D pose  can  be retrieved  using
RANSAC.   PVNet~\cite{Peng19}  chose  to  predict  the  directions  towards  the
reprojections,  rather than  a full  2D displacement,  but relies  on a  similar
RANSAC procedure  to estimate the  final 3D pose. \cite{Hu19}  and \cite{Peng19}
also  predict  the  masks of  the  target  objects,  in  order to  consider  the
predictions from the  image locations that lie  on the objects, as  they are the
only informative ones.

\subsection{DPOD and Pix2Pose}

\begin{figure}
\begin{center}
\includegraphics[width=0.8\linewidth]{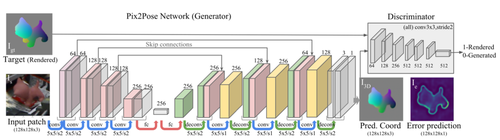}
\end{center}
\caption{\label{fig:pix2pose} Pix2Pose~\protect\cite{Park19} predicts the object
  coordinates of  the pixels lying  on the object  surface, and computes  the 3D
  pose from  these correspondences.  It relies  on a  GAN~\protect\cite{Goodfellow14} to
  perform   this  prediction   robustly  even   under  occlusion.    Image  from
  \protect\cite{Park19}.}
\end{figure}

DPOD~\cite{Zakharov19} and Pix2Pose~\cite{Park19} are also two methods that have
been presented  at the same conference  and present similarities. They  learn to
predict for  each pixel of  an input  image centered on  a target object  its 3D
coordinates  in the  object  coordinate  system, see  Figure~\ref{fig:pix2pose}.
More  exactly,  DPOD  predicts  the  pixels' texture  coordinates  but  this  is
fundamentally the same thing as  they both provide 2D-3D correspondences between
image locations and  their 3D object coordinates.  Such  representation is often
called 'object coordinates'  and more recently 'location field'  and has already
been used  for 3D  pose estimation in~\cite{Brachmann14,Brachmann16}  and before
that   in~\cite{Taylor12}  for   human  pose   estimation.   From   these  2D-3D
correspondences, it  is then possible to  estimate the object pose  using RANSAC
and P$n$P.   \cite{Park19} relies on a  GAN~\protect\cite{Goodfellow14} to learn
to perform  this prediction robustly even  under occlusion.  Note that  DPOD has
been demonstrated to run in real-time on a tablet.

Pix2Pose~\cite{Park19} also  uses a  loss function  that deals  with symmetrical
objects and can be written:
\begin{equation}
  \calL = \min_{\bR\in\text{Sym}} \frac{1}{|\calV|} \sum_{\bM \in \calV}{\| \RigidMotion(\bM; \hat{\pose}) - \RigidMotion(\bM; \bR.\bar{\pose}) \|_2} \> ,
  \label{eq:loss_sym}
\end{equation}
where  $\hat{\pose}$ and  $\bar{\pose}$ denote  the predicted  and ground  truth
poses  respectively, $\text{Sym}$  is a  set of  rotations corresponding  to the
object symmetries,  and $\bR.\pose$ denotes  the composition of such  a rotation
and a 3D pose. This deals with symmetries in a much more satisfying way than the
ADD-S loss~(Eq.~\eqref{eq:ADI}).

\subsection{Discussion}
As can be seen,  the last recent years have seen rich  developments in 3D object
pose estimation,  and methods have became  even more robust, more  accurate, and
faster.  Many methods rely on 2D segmentation to detect the objects, which seems
pretty robust, but assumes that only several  instances of the same object do not
overlap in  the image. Most  methods also rely on  a refinement step,  which may
slow things down, but relax the need for having a single strong detection stage.

There are still  several caveats though. First, it remains  difficult to compare
methods based on Deep Learning, even  though the use of public benchmarks helped
to promote fair comparisons.  Quantitative results depend not only on the method
itself, but also on how much effort  the authors put into training their methods
and augmenting  training data. Also,  the focus on  the benchmarks may  make the
method overfit  to their  data, and  it is  not clear  how well  current methods
generalize to  the real  world.  Also,  these methods rely  on large  numbers of
registered training images  and/or on textured models of the  objects, which can
be cumbersome to acquire.

%% file: category_pose.tex
So far,  we discussed  methods that  estimate the 3D  pose of  specific objects,
which  are  known in  advance.   As  we  already  mentioned, this  requires  the
cumbersome  step of  capturing  a training  set for  each  object. One  possible
direction to avoid  this step is to consider a  ``category''-level approach, for
objects  that  belong to  a  clear  category, such  as  'car'  or 'chair'.   The
annotation burden  moves then to images  of objects from the  target categories,
but  we  can then  estimate  the  3D pose  of  new,  unseen objects  from  these
categories.

Some early category-level methods only  estimate 3 degrees-of-freedom~(2 for the
image location and 1 for the rotation  over the ground plane) of the object pose
using regression,  classification or hybrid  variants of the two.   For example,
\cite{Xiang2016objectnet3d} directly  regresses azimuth, elevation  and in-plane
rotation     using    a     Convolutional     Neural    Network~(CNN),     while
\cite{tulsiani2015pose,tulsiani2015viewpoints} perform  viewpoint classification
by discretizing  the range  of each  angle into  a number  of disjoint  bins and
predicting the most likely bin using a CNN.

To estimate a full 3D pose, many methods rely on 'semantic keypoints', which can
be  detected in  the images,  and correspond  to 3D  points on  the object.  For
example, to estimate the  3D pose of a car, one may one  to consider the corners
of the roof  and the lights as semantic  keypoints.  \cite{pepik20153d} recovers
the pose from  keypoint predictions and CAD models using  a P$n$P algorithm, and
\cite{Pavlakos17}  predicts semantic  keypoints  and trains  a deformable  shape
model  which  takes  keypoint  uncertainties  into  account.   Relying  on  such
keypoints, however, is even more depending  in terms of annotations as keypoints
have to be  careful chosen and manually  located in many images, and  as they do
not generalize across categories.

\begin{figure}
  \begin{center}
    \includegraphics[width=0.8\linewidth]{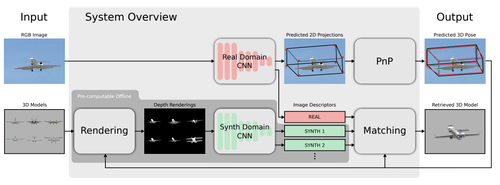}
  \end{center}
  \caption{\label{fig:grabner18} Estimating the  3D pose of an unknown  object from a
    known category. \protect\cite{Grabner18} predicts  the 2D reprojections of the
    corners of the  3D bounding box \emph{and}  the size of the  bounding box. For
    this information, it is possible to compute the pose of the object using a PnP
    algorithm. Image from \protect\cite{Grabner18}.}
\end{figure}

In fact, if one is not careful, the  concept of 3D pose for object categories is
ill-defined, as  different objects  from the  same category  are likely  to have
different sizes.  To properly define  this pose, \cite{Grabner18} considers that
the 3D pose  of an object from  a category is defined  as the 3D pose  of its 3D
bounding    box.     It    then     proposes    a    method    illustrated    in
Figure~\ref{fig:grabner18} that  extends BB8~\cite{Rad17}.   BB8 estimates  a 3D
pose  by predicting  the 2D  reprojections  of the  corners of  its 3D  bounding
box. \cite{Grabner18} predicts similar 2D reprojections, plus the size of the 3D
bounding box  in the form  of 3 values~(length,  height, width).  Using  these 3
values,  it is  possible to  compute  the 3D  coordinates  of the  corners in  a
coordinate system  related to the object.   From these 3D coordinates,  and from
the predicted 2D reprojections, it is possible  to compute the 3D pose of the 3D
bounding box.

\paragraph{3D model retrieval. } Since objects from the same category have various 3D models,
it may also be interesting to recover  a 3D model for these objects, in addition
to their  3D poses.   Different approaches  are possible. One  is to  recover an
existing 3D model from  a database that fits well the  object. Large datasets of
light  3D  models   exist  for  many  categories,  which   makes  this  approach
attractive~\cite{Chang15}.  To make this approach  efficient, it is best to rely
on metric learning, so that an embedding  computed for the object from the input
image    can    be    matched    against   the    embeddings    for    the    3D
models~\cite{Aubry2015understanding,Izadinia2017im2cad,Grabner18}.     If    the
similarity between  these embeddings can  be estimated based on  their Euclidean
distance  or  their dot  product,  then  it is  possible  to  rely on  efficient
techniques to perform this match.

\begin{figure}
  \begin{center}
    \begin{tabular}{cc}
      \includegraphics[width=0.45\linewidth]{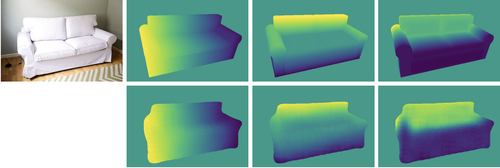} &
      \includegraphics[width=0.45\linewidth]{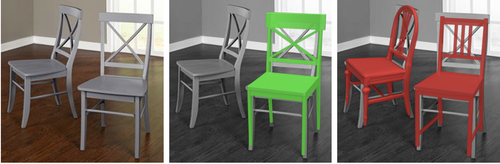} \\
    \end{tabular}

  \end{center}
  \caption{\label{fig:grabner19}  Recovering a  3D model  for an  unknown object
    from  a  known  category   using  an  image.   (a)  \protect\cite{Grabner19}
    generates renderings  of the  object coordinates~(called location  fields in
    the paper)  from 3D  models (top), and  predicts similar  object coordinates
    from the input images (bottom). Images  of objects and 3D models are matched
    based on embeddings computed from the  object coordinates. (b) An example of
    recovered 3D poses and  3D models for two chairs. The  3D models capture the
    general shapes of the real chairs, but do not fit perfectly.}
\end{figure}

The challenge is  that the object images  and the 3D models  have very different
natures, while  this approach  needs to compute  embeddings that  are comparable
from these two sources.  The solution is then to replace the  3D models by image
renderings; the  embeddings for the image  renderings can then be  compared more
easily with the embeddings for the  input images. Remains the domain gap between
the real input  images and the synthetic  image renderings. In~\cite{Grabner19},
as shown in Figure~\ref{fig:grabner19}, this  is solved by predicting the object
coordinates  for the  input image  using a  deep network,  and by  rendering the
object   coordinates  for   the  synthetic   images,  rather   than  a   regular
rendering.  This brings  the representations  for the  input images  and the  3D
models even  closer. From these  representations, it  is then easier  to compute
suitable embeddings.

Another approach  is to  predict the  object 3D  geometry directly  from images.
This  approach learns  a mapping  from the  appearance of  an object  to its  3D
geometry.  This  is more appealing  than recovering a  3D model from  a database
that has no guarantee to fit perfectly  the object, since this latter method can
potentially         adapt         better         to         the         object's
geometry~\cite{Shin18,Groueix18,Park19b}.  This is however much more challenging,
and  current  methods  are  still  often limited  to  clean  images  with  blank
background,  and sometimes  even synthetic  renderings, however  this is  a very
active area, and more progress can be expected in the near future.

%% file: hand_pose.tex
\section{3D Hand Pose Estimation from Depth Maps}

While related  to 3D object  pose estimation, hand  pose estimation has  its own
specificities.  On one  hand (no pun intended), it is  more challenging, if only
because  more degrees-of-freedom  must be  estimated. On  the other  hand, while
considering new 3D objects still requires  acquiring new data and/or specific 3D
models, a  single model can generalize  to unseen hands, despite  differences in
size, shape, and skin color, as these differences remain small, and considerably
smaller than differences between two  arbitrary objects. In practice, this means
that a model does not need new data  nor retraining to adapt to new users, which
is very convenient.

However, the motivation for 3D hand pose in AR applications is mostly for making
possible natural interaction  with virtual objects.  As we  will briefly discuss
in  Section~\ref{sec:manipulation}, convincing  interaction  requires very  high
accuracy, which current developments are aiming to achieve.

We will start  with 3D hand pose estimation  from a depth map. As  for 3D object
pose estimation,  the literature is  extremly large, and  we will focus  here as
well on a few representative methods.   This is the direction currently taken in
the industry,  because it  can provide  a robust and  accurate estimate.   It is
currently being deployed on commercial solutions, such as HoloLens 2 and Oculus.
We will  then turn  to hand  pose estimation  from color  images, which  is more
challenging, but  has also the potential  to avoid the drawbacks  of using depth
cameras. We will turn to pose estimation from color images in the next section.

\subsection{DeepPrior++}
Estimating the 3D pose of a hand from a depth map is in fact now relatively easy
and robust, when  using a well engineered solution based  on Deep Learning.  For
example,  \cite{Oberweger18} discusses  the  implementation of  a method  called
DeepPrior++, which is simple in principle and performs well.

DeepPrior++ first selects a  bounding box on the hand, and  then predicts the 3D
joint  locations from  the  depth data  within this  bounding  box. An  accurate
localization of  the bounding  appears to  improve the  final accuracy,  and the
method from the  center of mass of  the depth data within a  threshold, and then
uses a refinement step that can  be iterated.  This refinement step is performed
by  a regression  network predicting,  from a  3D bounding  box centered  on the
center of  mass, the 3D  location of  one of the  joints used as  a referential.
Then, the 3D joint locations can be predicted using a second Network that is not
very different from the one  presented in Figure~\ref{fig:basic_dn}, except that
it relies on  ResNet block~\cite{He16} for more accurate results.   The input to
this network is therefore the input depth  map cropped to contain only the hand,
and its output is made of the 3D coordinates of all the hand joints.

Simple data augmentation appears to improve accuracy, even when starting from an
already  large dataset.  \cite{Oberweger18} applies  random in-plane  rotations,
scales, 3D offsets to the input data.

\subsection{V2V-PoseNet}
V2V-PoseNet~\cite{Moon18} is a method that  has been identified by \cite{Yuan17}
as one  of the best performing  methods at the time,  based on the results  on a
public benchmark. \cite{Moon18}  argues that using a cropped depth  map as input
to   the   network  makes   training   difficult,   as   the  same   hand   pose
appears~(slightly) differently in a depth  map depending on where it reprojects.

\begin{figure}
  \begin{center}
    \includegraphics[width=0.95\linewidth]{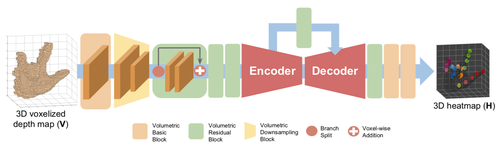}
  \end{center}
  \caption{\label{fig:V2V}  Architecture  of  V2V-PoseNet.  The  network  takes  a
    voxel-based representation of the depth map centered on the hand, and predicts
    the 3D joint locations in the form of heatmaps. }
\end{figure}

To avoid distortion, V2V-PoseNet converts the  depth map information in the hand
bounding  box into  a  voxel-based  representation (it  uses  a refinement  step
similar to  the one  in \cite{Oberweger18},  so accuracy  of the  hand detection
seems important). The  conversion is done by voxelizing the  3D bounding box for
the hand, and each voxel that contains at  least one pixel from the depth map is
set  to occupied.  The  voxel-based representation  therefore  corresponds to  a
dicretization of the hand surface.

Then, a network is trained to take this voxel-based representation as input, and
predicts the 3D joint  locations in the form of 3D heatmaps.  A  3D heatmap is a
3D array containing  confidence values. There is one 3D  heatmap for each joint,
and the joint location is taken as the 3D location with the higher confidence.

\subsection{A2J}
\begin{figure}
  \begin{center}
    \begin{tabular}{cc}
      \includegraphics[width=0.45\linewidth]{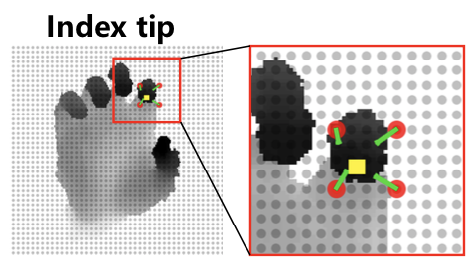}&
      \includegraphics[width=0.45\linewidth]{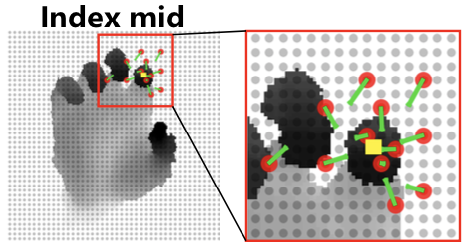}\\
    \end{tabular}
  \end{center}
  \caption{\label{fig:A2J}  The  A2J  method  predicts,  for  regularly  sampled
    ``anchor points'' (gray dots), and for each joint, a 2D displacement towards
    the 2D joint  location (green line segments), its depth,  and the confidence
    for  these predictions.  Red  dots  correspond to  anchor  points with  high
    confidences. The 3D joint locations can  be estimated from these as weighted
    sums~(yellow squares). Images from \protect\cite{Xiong19}.}
\end{figure}

The A2J method~\cite{Xiong19} performs similarly or better than V2V-PoseNet, but
faster.  The  prediction of  the 3D  joint locations is  performed via  what are
called in the paper ``anchor points''.  Anchor points are 2D locations regularly
sampled over  the bounding  box of  the hand,  obtained as  in~\cite{Moon18} and
\cite{Oberweger18}.  As shown  in Figure~\ref{fig:A2J}, a network  is trained to
predict, for each  anchor point, and for  each joint, 1) a  2D displacement from
the anchor  point to  the joint,  2) the  depth of  the joint,  and 3)  a weight
reflecting  the   confidence  for   the  prediction.    The  prediction   of  2D
displacements makes  this method  close to \cite{Hu19}  and PVNet~\cite{Peng19},
which were developed for 3D object pose estimation under occlusions.

From  the output  of  the network,  it  is  possible to  estimate  the 3D  joint
locations as weighted sums of the 2D  locations and depths, where the weights of
the sums are  the ones predicted by  the network. Special care is  taken to make
sure the weights sum to 1. The  paper argues that this is similar to predictions
by an ``ensemble'',  which are known to  generalize well~\cite{Breiman96}.  The
weights play a  critical role here: Without them, the  predicted locations would
be a linear combination of the network output, which could be done by a standard
layer without  having to introduce  anchor points.  The product  between weights
and  2D displacements  and  depths make  the predicted  joint  locations a  more
complex, non-linear transformation of the network output.

\subsection{Discussion}
As  discussed  in detail  in~\cite{Armagan20},  the  coverage  of the  space  of
possible hand poses is critical to  achieve good performance. This should not be
surprising: Deep  Networks essentially learn  a mapping between the  input (here
the depth map) and the output (here  the hand pose) from the training set.  They
can interpolate very well between samples,  but poorly extrapolate: If the input
contains a pose too different from any sample in the training set, the output is
likely to be  wrong.  When using as  input depth maps, which do  not suffer from
light changes or cluttered background like  color images, having a good training
set  is  the main  important  issue.

That was in  fact already discussed in \cite{Oberweger18}, and  even before that
in \cite{Supancic15}.  Combining  the DeepPrior++ that relies on  a simple mapping
with  a  simple domain  adaptation  method~\cite{Rad18}  in  order to  use  more
synthetic data appears  to perform very well, and still  outperfoms more complex
methods on the  NYU dataset, for example.   While it is not  entirely clear, the
direction taken by HoloLens  2 and Oculus is to train a simple  model on a large
set  of synthetic  training data,  and  refine the  pose estimate  using an  ICP
algorithm~\cite{Sharp15} aligning a 3D model of the hand to the depth data.

\section{3D Hand Pose Estimation from an RGB Image}

We now turn to  modern approaches to 3D hand pose estimation  from an RGB image,
which is significantly  more challenging than from  a depth map, but  which is a
field that has been impressively fast progress over the past few years.

\begin{figure}
\begin{center}
\includegraphics[width=0.8\linewidth]{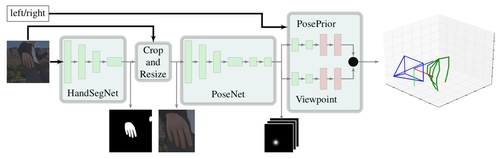}
\end{center}
\caption{\label{fig:zimmerman17} Overview of \protect\cite{Zimmerman17}. The hand is first
  segmented in 2D to obtain a 2D bounding box. From the image information within
this bounding box, a set of heatmaps for the hand joints is predicted and lifted
to 3D. Image from \protect\cite{Zimmerman17}.}
\end{figure}

\subsection{\protect\cite{Zimmerman17}}

\cite{Zimmerman17} was one of the first methods predicting the 3D hand pose from
a  single color  image, without  using depth  information.  An  overview of  the
method is shown in Figure~\ref{fig:zimmerman17}.  The hand is first segmented in
2D,  as in  recent approaches  for 3D  object pose  estimation, to  obtain a  2D
bounding box centered  on it.  From this  window, a 2D heatmap  is predicted for
each  joint,  using  Convolutional Pose  Machines~\cite{Wei16}.   This  provides
information on the 2D joint locations. To obtain the final 3D pose, the heatmaps
are  lifted in  3D  by using  2  networks:  One network  predicts  the 3D  joint
locations in a canonical system attached to the hand; the other network predicts
the 3D pose (3D  rotation and translation) of the hand  in the camera coordinate
system.  From these 2 set of information, it is possible to compute the 3D joint
coordinates in  the camera  coordinate system.  This is  shown to  work slightly
better than a  single network directly predicting the 3D  joint locations in the
camera coordinate system.

To  train  the  segmentation,  heatmap  prediction,  and  3D  pose  predictions,
\cite{Zimmerman17} created  a dataset of  synthetic images  of hands, as  the 3D
pose can  be known exactly. To  do so, they  used freely available 3D  models of
humans  with corresponding  animations  from Mixamo,  a  company specialized  in
character   animation,  and   Blender   to  render   the   images  over   random
background. Lighting, viewpoint,  and skin properties were  also randomized. The
dataset is publicly available online.

\subsection{\protect\cite{Iqbal18}}
Like \cite{Zimmerman17}, \cite{Iqbal18} relies on 2D heatmaps from a color image
to lift them to 3D, but with several differences that improve the pose accuracy.
They oppose two  approaches to 3D pose prediction: The  first approach relies on
heatmaps; it is generally accurate but keeping this accuracy with 3D heatmaps is
still intractable, as a  finely sampled 3D volume can be  very large. The second
approach is  to predict the  3D joints in a  ``holistic'' way, meaning  that the
values of the 3D joint locations are directly predicted from the input image, as
was  done   in~\cite{Toshev14,Sun17}  for   human  body  pose   prediction,  and
\cite{Oberweger17} for hand pose prediction from depth data. Unfortunately,
``holistic'' approaches tend to be less accurate.

\begin{figure}
\begin{center}
\includegraphics[width=\linewidth]{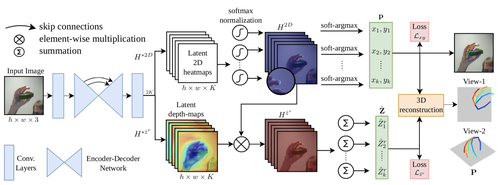}
\end{center}
\caption{\label{fig:iqbal18} Architecture of \protect\cite{Iqbal18}. The method
  predicts a heatmap and a depth map for each joint, and estimates the 3D pose
  using geometric constraints. The model is trained end-to-end, which lets the
  model find optimal heatmaps and depth maps. Image from \protect\cite{Iqbal18}.}
\end{figure}

To  keep the  accuracy of  heatmaps,  and to  predict  the joint  depths with  a
tractable method, \cite{Iqbal18} also predicts,  in addition to the 2D heatmaps,
a depth map for each joint, where  the depth values are predicted depths for the
corresponding  joint   relative  to  a   reference  joint.  This  is   shown  in
Figure~\ref{fig:iqbal18}.  The  final  3D   pose  is  computed  using  geometric
constraints from this information.  Instead  of learning to predict the heatmaps
in a supervised way, \cite{Iqbal18} learns to predict ``latent'' heatmaps.  This
means that the deep model is pretrained  to predict heatmaps and depth maps, but
the constraints  on the predicted heatmaps  and depth maps are  removed from the
full loss, and  the model is trained ``end-to-end'', with  a loss involving only
the final predicted 3D pose.  This lets the optimization find heatmaps and depth
maps that are easier to predict, while  being more useful to predict the correct
pose.

\subsection{GANerated Hands}
As  already mentioned  above,  training data  and its  diversity  is of  crucial
importance in  Deep Learning.  For  3D hand  pose estimation from  color images,
this means that training  data need to span the hand pose  space, the hand shape
space, possible lighting,  possible background, etc. Moreover, this  data has to
be annotated in  3D, which is very challenging.  Generating  synthetic images is
thus attractive,  but because of the  ``domain gap'' between real  and synthetic
images, this may impact the accuracy of  a model trained on synthetic images and
applied to real images.

To bridge  the domain  gap between synthetic  and real  images, \cite{Mueller18}
extends  CycleGAN~\cite{Zhu17},  which  is  itself based  on  Generative  Adversarial
Networks~(GANs).  In original GANs~\cite{Goodfellow14}, a network is trained to
generate images  from random vectors.  In  order to ensure the  generated images
are ``realistic'',  a second network  is trained jointly  with the first  one to
distinguish the  generated images from real  ones.  When this network  fails and
cannot classify the images generated by the first network as generated, it means
these generated  images are  similar to real  ones.  CycleGAN~\cite{Zhu17}  builds on
this idea  to change an image  from a domain  into an image with  the \emph{same
  content} but similar to the images from another domain.  An image generator is
thus train to take an image from some  domain as input and to generate as output
another image.   A second network ensures  that the output looks  similar to the
target domain. To make sure the content of the image is preserved, a second pair
of networks is jointly trained with the  first one: This pair is trained jointly
with  the first  one, to  re-generate the  original input  image from  the image
generated by the first generator.

\begin{figure}
\begin{center}
\includegraphics[width=0.6\linewidth]{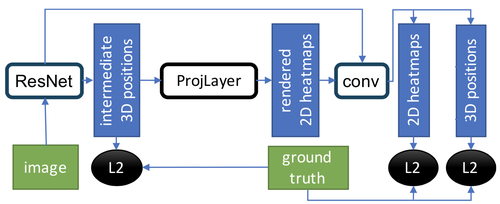}
\end{center}
\caption{\label{fig:GANerated}       The        architecture       used       in
  GANerated~\protect\cite{Mueller18}  predicts both  2D heatmaps  and 3D  joints
  locations,   and   estimates  the   3D   hand   pose  with   a   (non-learned)
  optimization. Image from \protect\cite{Mueller18}.}
\end{figure}

\cite{Mueller18} extends  CycleGAN by adding  a constraint between  the original
image  (here   a  synthetic  image)  and   the  image  returned  by   the  first
generator. The synthetic image is a hand rendered over a uniform background, and
a segmentation  network is trained  jointly to  segment the generated  image, so
that the  predicted segmentation  matches the  segmentation of  the hand  in the
synthetic image.  This helps preserving, at least to some extent, the 3D pose of
the  hand  in  the  transformed  image.   Figure~\ref{fig:GANerated}  shows  two
synthetic  images transformed  by CycleGAN,  with and  without this  constraint.
Another  advantage is  that the  plain  background can  be changed  by a  random
background to augment the dataset of transformed images.

Besides this data generation, the 3D hand pose prediction in \cite{Mueller18} is
also interesting.  A network is trained  to predict \emph{both} 2D  heatmaps for
the 2D  joint locations, and  the 3D joint locations.  Given the output  of this
network over  a video stream,  an optimization is  performed to fit  a kinematic
model to these 2D and 3D  predictions under joint angle constraints and temporal
smoothness constraints. The output of the network is therefore not used directly
as the 3D hand pose, but used as a observation in this final optimization.

\begin{figure}
\begin{center}
\includegraphics[width=\linewidth]{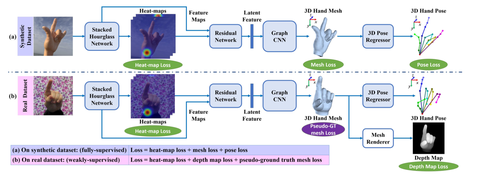}
\end{center}
\caption{\label{fig:Ge19} The architecture used in \protect\cite{Ge19} predicts
  2D heatmaps for the joints and a 3D mesh for the hand. Image from \protect\cite{Ge19}.}
\end{figure}

\subsection{3D Hand Shape and Pose Estimation~\protect\cite{Ge19,Boukhayma19}}

\cite{Ge19}    and   \cite{Boukhayma19}    were    published    at   the    same
conference~(CVPR'19) and both propose a method  for predicting the hand shape in
addition to  the hand pose.

\cite{Ge19} first generates  a training set of hands under  various poses, where
each image is annotated with the pose and shape parameters, with special care to
make the  rendered images as  realistic as  possible.  Using this  dataset, they
train a network~(shown in Figure~\ref{fig:Ge19}) to predict  a ``latent feature
vector'', which is  fed to a Graph CNN~\cite{Defferrard16}.   The motivation for
using a Graph CNN is to predict a 3D mesh for the hand, which can be represented
as a graph.  The loss function thus a loss that compares, for synthetic training
images, the mesh  predicted by the GraphCNN  and the ground truth  mesh, and the
predicted 3D pose and the ground truth  pose. For real training images, which do
not have ground  truth mesh available, they  use the pose loss term,  but also a
term that compares  the predicted 3D mesh  with the depth map for  the input RGB
image when available, and another term  that compares the same predicted 3D mesh
with a mesh predicted from the ground truth 2D heatmaps.

\begin{figure}
\begin{center}
\includegraphics[width=\linewidth]{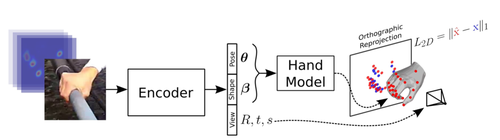}
\end{center}
\caption{\label{fig:Boukhayma19}  \protect\cite{Boukhayma19}  learns to  predict
  the pose and  shape parameters of the  hand MANO model~\protect\cite{Romero17}
  without  3D  annotations  on  the  pose   and  shape,  by  minimizing  on  the
  reprojection error of the joints. Image from \protect\cite{Boukhayma19}.}
\end{figure}

The  method   of  \cite{Boukhayma19}  is   simpler.   It  relies  on   the  MANO
model~\cite{Romero17}, which is a popular  deformable and parameterized model of
the  human hand,  created from  many  captures of  real hands.   The MANO  model
provides a  differentiable function that generates  a 3D model of  a hand, given
pose parameters  and shape parameters. It  does not require a  dataset annotated
with  the  shape parameters,  but  instead  it  can re-uses  existing  datasets,
annotated only with the 3D hand pose.  To  do so, they train a network (shown in
Figure~\ref{fig:Boukhayma19}) to predict the hand shape and pose parameters, and
the  loss function  is the  2D  distances between  the reprojections  of the  3D
joints,  as computed  from  the  predicted shape  and  pose  parameters and  the
annotated 2D joints.

\subsection{Implementation in MediaPipe}
We can also  mention a Google Project~\cite{Bazarevsky19} for  real-time 3D hand
pose implementation on  mobile devices. They first detect the  palm, rather than
the full hand: This allows them to  only consider squared bounding boxes, and to
avoid confusion between  multiple hand detections. The 3D hand  pose is directly
predicted as the 3D joint locations, using a model trained on real and synthetic
images.

Unfortunately, the work  is not presented in a formal  research publication, and
it  is difficult  to  compare their  results  with the  ones  obtained by  other
methods.  However, the implementation is publicly available, and appears to work
well.  This may  show again  that good  engineering can  make simple  approaches
perform well.

\subsection{Manipulating Virtual Objects}
\label{sec:manipulation}
\input{manipulation}

%% file: manipulation.tex
One of the main motivations  to track the 3D pose of a hand  for an AR system is
to  make possible  natural interaction  with virtual  objects.  It  is therefore
important to be  able to exploit the  estimated 3D hand pose to  compute how the
virtual objects should  move when they are in interaction  with the user's hand.

This is not a trivial problem.  In  the real world, our ability to interact with
objects is due to  the presence of friction between the  surfaces of the objects
and our hands, as friction is a force that resists motion.  However, friction is
very challenging to simulate correctly.  Also, nothing prevents the real hand to
penetrate the virtual objects, and make realism fail.

Some approaches rely  on heuristics to perform a set  of object interactions. It
may be reasonable to focus on object grasping  for AR systems, as it is the most
interesting   interaction.   Some   methods  are   data-driven  and   synthesize
prerecorded, real hand data to identify the most similar one during runtime from
a                                                                     predefined
database~\cite{li2007data,miller2003automatic,rijpkema1991computer}.

A friction model from physics,  the \textit{Coulomb-Contensou} model was used in
\cite{talvas2015aggregate} to reach more accurate results, but the computational
complexity  becomes very  high, and  not necessarily  compatible with  real-time
constraints.  \cite{Hoell18} relies on the simpler Coulomb model that appears to
be a good trade-off between realism  and tractability.  The force applied by the
user are taken  proportional to how much their real  hand penetrates the virtual
objects.  Computing accurately the forces  requires very accurate 3D hand poses,
otherwise they can become very unstable.

Very recently, videos  demonstrating the real-time interaction  system of Oculus
using depth-based  hand tracking  appeared. No  technical details  are available
yet, but the system seems to allow for realistic grasping-based interactions.

%% file: object_hand_pose.tex
We finally turn  to the problem of estimating  a 3D pose for both a  hand and an
object, when  the hand directly  manipulates the  object.  Even if  the existing
methods  are  still  preliminary,  the  potential  application  to  AR  is  very
appealing, as this  could offer tangible interfaces by  manipulating objects, or
even the possibility to bring real objects into the virtual world.

This problem still remains very challenging, as as the close contact between the
hand and  the object results  in mutual occlusions  that can be  large.  Dealing
with egocentric  views, which are  more relevant  for AR applications,  is often
even more challenging.  Fortunately, there are also physical constraints between
the hand and  the object, such as impossibility of  penetration but also natural
grasping poses, that may help solving this problem.

Pioneered          approaches           joint          hand+object          pose
estimation~\cite{Oikonomidis2011full,Wang11d,Ballan2012motion}  typically relied
on      frame-by-frame      tracking,      and     in      the      case      of
\cite{kyriazis2013physically,Tzionas2016},  also  on   a  physics  simulator  to
exploit  the  constraints  between  hand and  object.   However,  frame-to-frame
tracking, when used alone, may require careful initialization from the user, and
may drift over time.   The ability to estimate the 3D poses  from a single image
without prior from previous frames or  from the user makes tracking more robust.
Given the difficulty  of the task, this  has been possible only with  the use of
Deep Learning.

\subsection{ObMan and HOPS-Net}

ObMan~\cite{Hasson19}      and     HOPS-Net~\cite{kokic2019learning}      almost
simultaneously proposed  methods with  some similarities  for estimating  the 3D
pose \emph{and  shape} of both a  hand and of  the object it manipulates  from a
single RGB image.  Creating realistic training  data is the first challenge, and
they both created a  dataset of synthetic images of hands  and objects using the
\textit{GraspIt!}    simulation   environment~\cite{Miller04}   for   generating
realistic grasps given 3D models for the hand and for the object.

\cite{Hasson19} train their model on  their synthetic dataset before fine-tuning
it on \cite{Garcia2018first}, which provide  annotated real images.  To make the
synthetic  images  more   realistic,  \cite{kokic2019learning}  use  ``Augmented
CycleGAN''~\cite{Almahairi18}: The  motivation to use Augmented  CycleGAN rather
than CycleGAN is that the former  can deal with many-to-many mappings, while the
latter  considers only  one-to-one  mappings,  and that  a  synthetic image  can
correspond to multiple real images.

\begin{figure}
  \begin{center}
    \includegraphics[width=0.7\linewidth]{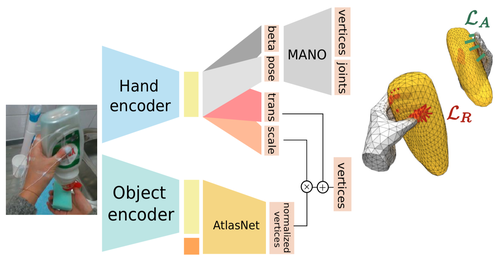}
  \end{center}
  \caption{\label{fig:hasson19} \protect\cite{Hasson19} learns to directly predict
    pose and shape parameters for the hand, and shape for the object, using a term
    that  encourages   the  hand  and  the   object  to  be  in   contact  without
    interpenetration. Image from \protect\cite{Hasson19}.}
\end{figure}

As shown  in Figure~\ref{fig:hasson19}, \cite{Hasson19}  does not go  through an
explicit 2D  detection of  the object or  the hand, by  contrast with  the other
methods we  already discussed.  Instead,  the method directly predicts  from the
input image the pose  and shape parameters for the hand  (using the MANO model),
and  the  shape  for  the  object.   To  predict  the  object  shape,  they  use
AtlasNet~\cite{Groueix18}, a method  that can predict a set of  3D points from a
color  image. This  shape prediction  does  not require  to know  the object  in
advance,  and predicts  the shape  in the  camera coordinates  system, there  is
therefore no notion of object pose.  The loss function to train the architecture
includes a term that prevents intersection  between the hand and the object, and
a term  that penalizes cases in  which the hand is  close to the surface  of the
object without being in contact.

\begin{figure}
  \begin{center}
    \includegraphics[width=\linewidth]{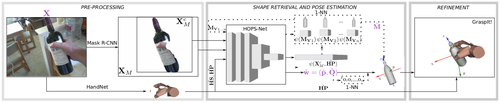}
  \end{center}
  \caption{\label{fig:kokic19} \protect\cite{kokic2019learning} first predicts the
    hand pose and  detects the object in  2D, then predicts the  object's pose for
    its 2D  bounding pose  and the hand  pose. It  also finds a  3D model  for the
    object from its appearance, and uses it  to refine the object pose so that the
    object     is     in     contact     with     the     hand.     Image     from
    \protect\cite{kokic2019learning}.}
\end{figure}

\cite{kokic2019learning}  focuses on  objects from  specific categories~(bottle,
mug, knife, and bowl), and uses Mask R-CNN~\cite{He17} to detect and segment the
object: The object  does not have to be  known as long as it belongs  to a known
category. The  method also relies on  \cite{Zimmerman17} to predict the  3D hand
pose.   A network  is trained  to predict,  from the  bounding box  predicted by
Mask-RCNN and the  hand pose (to provide additional constraints),  the 3D object
pose and a description vector for the shape.  The 3D pose representation depends
on the  category to handle symmetries  in the object shape,  and the description
vector is matched against a database of 3D  models to retrieve a 3D model with a
similar shape~(see \cite{Grabner19} discussed in Section~\ref{sec:category}).  A
final refinement step  optimizes the object pose  so that it is  in contact with
the hand.

\begin{figure}
  \begin{center}
    \includegraphics[width=\linewidth]{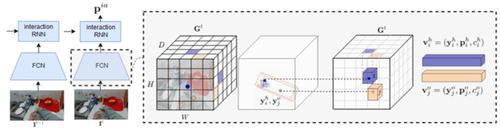}
  \end{center}
  \caption{\label{fig:H+O} The H+O method~\protect\cite{tekin2019h} predicts the
    hand  and object  poses both  as a  set of  3D locations,  to get  a unified
    representation  and simplify  the  loss function.   These  3D locations  are
    predicted \textit{via} a 3D grid.   Moreover, the predicted 3D locations are
    provided to a  recurrent neural network, to propagate  this information over
    time  and  recognize   the  action  performed  by  the   hand.   Image  from
    \protect\cite{tekin2019h}.}
\end{figure}

\subsection{H+O~\protect\cite{tekin2019h}}
H+O~\cite{tekin2019h} is a method that can predict the 3D poses of a hand and an
object from  an image,  but it  also learns  a temporal  model to  recognize the
performed actions (for example, pouring,  opening, or cleaning) and to introduce
temporal constraints to improve the  pose estimation. Moreover, the architecture
remains relatively simple,  as shown in Figure~\ref{fig:H+O}.  It  is trained on
the           First-Person           Hand          Action           dataset~(see
Section~\ref{sec:hand_object_datasets}).

The 3D pose of the object is predicted as the 3D locations of the corners of the
3D bounding box---from this information, it is possible to compute a 3D rotation
and translation, using  \cite{Gower75} for example.  Since the 3D  hand pose can
also be  predicted as  a set of  3D locations,  this makes the  2 output  of the
network consistent, and  avoids a weight to  tune between the loss  term for the
hand and the loss  term for the object.  More exactly, the  3D locations are not
predicted directly.   The 3D space  is split into cells,  and for each  cell and
each 3D location, it  is predicted if the 3D location is  within this cell, plus
an offset  between a  reference corner  of the cell  and the  3D location---this
offset ensures that  the 3D location can be predicted  accurately, even with the
space discretization  into cells.  A  confidence is  also predicted for  each 3D
location prediction. The 3D hand and object  poses are computed by taking the 3D
points with the highest confidences.

To  enforce both  temporal constraints  and  recognize the  actions, the  method
relies on  a Recurrent Neural  Network, and more  exactly on an  Long Short-Term
Memory~(LSTM)~\cite{Hochreiter97}.  Such network  can propagate information from
the predicted 3D locations over time, and  it is used here to predict the action
and the  nature of  the object,  in addition to  be a  way to  learn constraints
between the hand and the object.

\subsection{HOnnotate~\protect\cite{Hampali20}}

\cite{Hampali20}  proposed a  method to  automatically annotate  real images  of
hands      grasping     objects      with      their      3D     poses      (see
Figure~\ref{fig:hand_datasets}(f)), which  works with  a single  RGB-D camera,
but  can   exploit  more  cameras   if  available  for  better   robustness  and
accuracy. The main idea is to optimize jointly  all the 3D poses of the hand and
the object over the sequence to exploit temporal consistency. The method is
automated, which means that it can be used easily to labeled new sequences.

The authors use  the resulting dataset called  HO-3D to learn to  predict from a
single RGB image the  3D pose of a hand manipulating an object.  This is done by
training a Deep Network to predict the 2D joint locations of the hand along with
the joint direction vectors and lift them to 3D by fitting a MANO model to these
predictions.  This reaches good accuracy  despite occlusions by the object, even
when the object was not seen during training.

%% file: future.tex
This chapter aimed at demonstrating and explaining the impressive development of
3D pose estimation in computer vision since  the early pioneer works, and in the
context of  potential applications  to Augmented  Reality. Methods  are becoming
more robust and accurate, while benefiting  on fast implementations on GPUs.  We
focused on some popular works, but they  are only entries to many other works we
could not present here, and that the reader can explore on their own.

However, as  we pointed out at  the end of Section~\ref{sec:object_pose},  it is
still too early to draw conclusions on  what is the best methodology for 3D pose
estimation.   Quantitative results  not only  reflect the  contributions of  the
methods,  and  also  depend  on  how  much  effort  the  authors  put  in  their
implemetation.  As a result, there are sometimes contractory conclusions between
papers.  For example, is it important to  first detect the object or the hand in
2D first, or not?  Are 2D heatmaps  important for accuracy, or not? What is the
best pose representation for a 3D pose:  A 3D rotation and translation, or a set
of 3D points?  Also, performance is not the only aspect here: For example, using
a set  of 3D points  for both the object  and hand poses  as in~\cite{tekin2019}
relaxes for tuning  the weights in the loss function,  which is also interesting
in practice.

The  focus on  some  public benchmarks  may  also bias  the  conclusions on  the
performance of  the methods in  the real world: How  well do they  perform under
poor  lighting?   How  well  do   they  perform  on  different  cameras  without
retraining?

Another aspect that needs to be solved in the dependence on training sets, which
are complex to create, and to a training  stage each time a new object should be
considered.  The  problem does  not really  occur for hands,  as a  large enough
training  set for  hands  would ensure  gene ralization  to  unseen hands.   For
objects, variability  in terms of  shape and appearance  is of course  much more
important.   Some methods  have  shown some  generalization  power within  known
categories~\cite{Grabner19,kokic2019learning}.  \cite{Xiang2018}  has shown some
generalization to new objects, but only on simple synthetic images and for known
3D models. Being able to consider  instantly (without retraining nor capturing a
3D  model) any  object in  an AR  system is  a dream  that will  probably become
accessible in the next years.